\pdfoutput=1

\documentclass[11pt]{article}

\usepackage[preprint]{acl}

\usepackage{times}
\usepackage{latexsym}

\usepackage[T1]{fontenc}

\usepackage[utf8]{inputenc}

\usepackage{microtype}

\usepackage{inconsolata}

\usepackage{graphicx}
\usepackage{booktabs}
\usepackage{xcolor}
\usepackage{algorithm}
\usepackage{algpseudocode}
\usepackage{tcolorbox}
\usepackage{csquotes}
\tcbuselibrary{skins}

\usepackage{romannum}
\definecolor{cGrey}{RGB}{248, 248, 248}
\definecolor{cYellow}{RGB}{255,255,3}
\definecolor{cBlue}{RGB}{69,123,157}
\definecolor{cRed}{RGB}{231,56,71}
\definecolor{cRed_1}{RGB}{191,30,46}
\definecolor{cGray}{RGB}{168,218,219}
\definecolor{cBlue_2}{RGB}{5,48,97}
\definecolor{cBlue_1}{RGB}{115,186,214}
\definecolor{cBlue_3}{RGB}{13,76,109}
\definecolor{cBlue_4}{RGB}{64,121,160}
\definecolor{cOrange}{RGB}{250,134,0}
\definecolor{cBlue_6}{RGB}{13,76,109}
\definecolor{cBlue_7}{RGB}{16,106,130}
\definecolor{cBlue_8}{RGB}{19,136,160}
\newcommand*{\affaddr}[1]{#1} 
\newcommand*{\affmark}[1][*]{\textsuperscript{#1}}

\title{Towards Multi-Agent Reasoning Systems for Collaborative Expertise Delegation: An Exploratory Design Study}

\author{Baixuan Xu\affmark[1]\thanks{\quad Equal Contribution},
Chunyang Li\affmark[1]$^{*}$,
Weiqi Wang\affmark[1]$^{*}$,
Wei Fan\affmark[1],
Tianshi Zheng\affmark[1],
Haochen Shi\affmark[1],\\
\textbf{
Tao Fan\affmark[1]$^{,}$\affmark[2],
Yangqiu Song\affmark[1],
Qiang Yang\affmark[3]}\\
\affaddr{\affmark[1]The Hong Kong University of Science and Technology}
\affaddr{\affmark[2]WeBank}\\
\affaddr{\affmark[3]The Hong Kong Polytechnic University\\}
\texttt{\{bxuan, cliei, wwangbw\}@cse.ust.hk}\\}

\begin{document}
\maketitle
\pagenumbering{arabic}
\begin{abstract}

Designing effective collaboration structure for multi-agent LLM systems to enhance collective reasoning is crucial yet remains under-explored. 
In this paper, we systematically investigate how collaborative reasoning performance is affected by three key design dimensions: (1) Expertise-Domain Alignment, (2) Collaboration Paradigm (structured workflow vs. diversity-driven integration), and (3) System Scale. 
Our findings reveal that expertise alignment benefits are highly domain-contingent, proving most effective for contextual reasoning tasks. 
Furthermore, collaboration focused on integrating diverse knowledge consistently outperforms rigid task decomposition. 
Finally, we empirically explore the impact of scaling the multi-agent system with expertise specialization and study the computational trade off, highlighting the need for more efficient communication protocol design.
This work provides concrete guidelines for configuring specialized multi-agent system and identifies critical architectural trade-offs and bottlenecks for scalable multi-agent reasoning.
The code will be made available upon acceptance.

\end{abstract}

\section{Introduction}
Collective intelligence, the emergent problem-solving capability arising from structured group interactions, has long been recognized as a cornerstone of complex human decision-making~\cite{Surowiecki2004Wisdom}. 
Through mechanisms like deliberative debate and systematic knowledge integration, human collectives consistently outperform individual experts in tasks requiring multi-perspective analysis and contextual synthesis.

\begin{figure}[t]
    \centering
    \includegraphics[width=\linewidth]{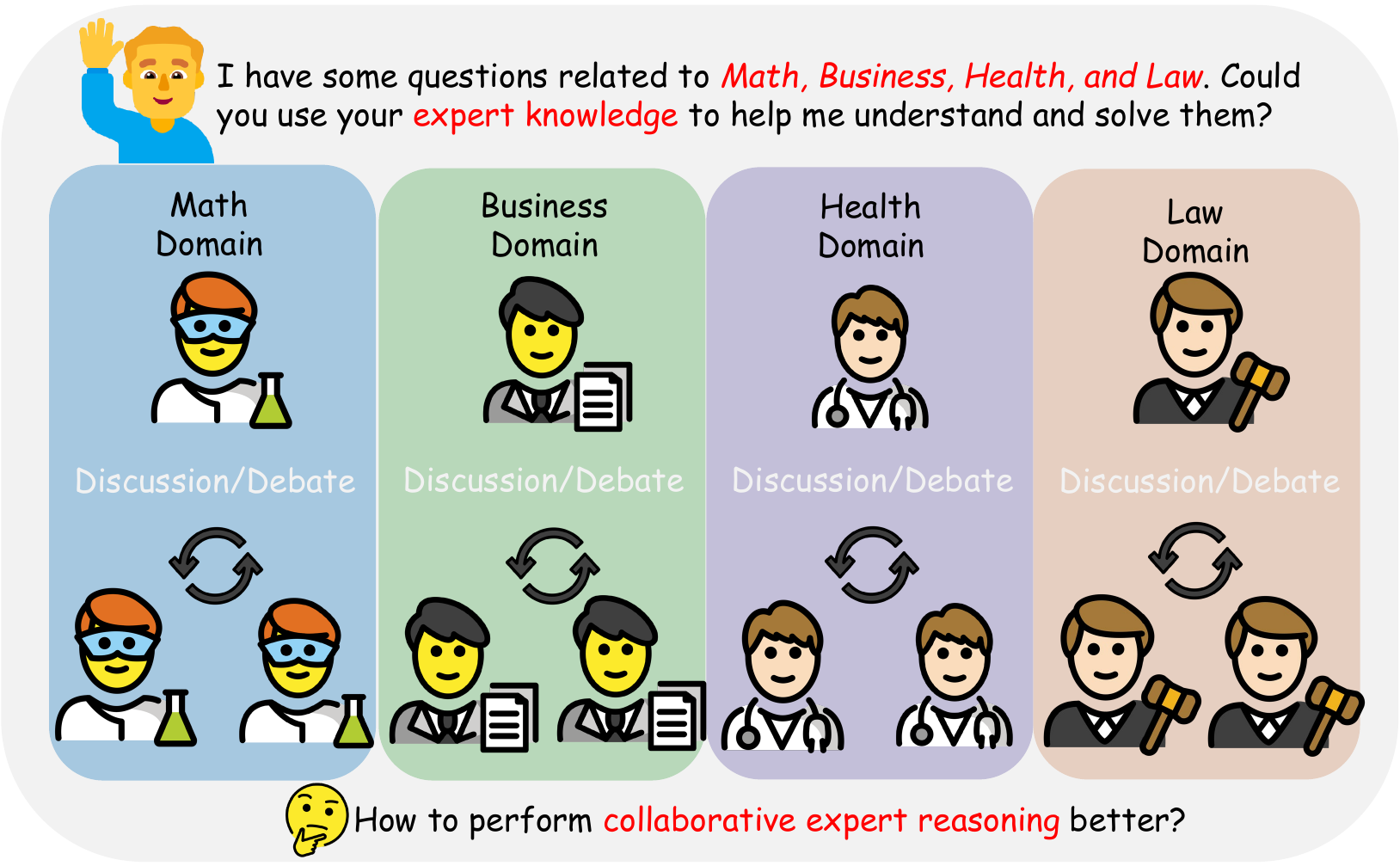}
    \caption{Workflow diagram for a multi-agent reasoning system with specialized agents.}
    \label{fig:intro}
    \vspace{-0.4cm}
\end{figure}
The recent evolution of large language and reasoning models (LLMs/LRMs)~\cite{DBLP:journals/corr/abs-2501-15383,DBLP:journals/corr/abs-2412-16720,DBLP:journals/corr/abs-2501-12599} has spurred parallel investigations into machine collective intelligence. 
Contemporary research has developed artificial analogs of human collaboration patterns through techniques such as multi-agent debate frameworks and workflow orchestration~\cite{DBLP:conf/emnlp/LiDZHGLI24,DBLP:conf/icml/Du00TM24,DBLP:conf/emnlp/Liang0JW00Y0T24}. 
These approaches primarily focus on either enhancing individual model performance through collective verification processes or establishing general-purpose problem-solving workflow pipelines.

A prevalent strategy to enhance collective intelligence in these systems is collaborative expertise specialization, where LLMs are instructed to simulate specific expert personas (e.g., ``act as an experienced lawyer'')~\cite{DBLP:journals/corr/abs-2411-03284,DBLP:conf/emnlp/Xu0S0JFBLYLLYYC24}. 
This approach is hypothesize to operate through two primary mechanisms: (1) Knowledge Recall, activating relevant domain-specific knowledge latent within the LLM via contextual role framing, and (2) Perspective Synthesis, leveraging diverse expert viewpoints to foster emergent, robust problem-solving patterns.

Although expertise specialization is widely adopted in multi-agent system~\cite{DBLP:journals/corr/abs-2406-04692,DBLP:journals/corr/abs-2411-03284}, the impact of varying collaborative expertise domain on distinct downstream scenarios remains underexplored.
To address this gap, we empirically evaluate the influence of different collaborative expertise configurations on task performance across four representative domains from MMLU-pro~\cite{DBLP:conf/nips/WangMZNCGRAHJLK24}.
\textbf{Our findings demonstrate a positive correlation between task performance and the alignment of group expertise with the task domain}, providing crucial insights for optimizing expertise specialization in multi-agent systems.


Another dimension concerning the multi-agent system design is the collaboration paradigm\textemdash namely, the mechanism governing how specialist agents interact in a multi-agent system. 
Currently, the collaboration paradigm predominantly used in recent studies could be categorized into two kinds: (1) Diversity-Driven Perspective Integration, where agents, often embodying different viewpoints or roles, are encouraged to generate diverse responses to enrich the solution space~\cite{DBLP:conf/acl/WangWSTS24,DBLP:conf/acl/ChenSB24,DBLP:conf/coling/HuCL025}. 
(2) Structured Workflow Cooperation, where different agents are assigned distinct sub-tasks within a predefined pipeline to collaboratively construct a solution~\cite{DBLP:conf/iclr/ChenSZ0YCYLHQQC24,DBLP:conf/iclr/HongZCZCWZWYLZR24}.
To understand the preference of collaboration paradigm in collaborative expertise specialization, we design comparative experiments which unveil the performance differences between paradigms.
\textbf{Our observations reveal a consistent advantage for diversity-driven collaboration over structured workflow collaboration, suggesting the superiority of the diversity-driven paradigm.}
Detailed analysis regarding viewpoint diversity are also conducted to study the impact of agent response diversity in multi-agent system, where a higher diversity could indicate a better performance.

Finally, constructing large-scale multi-agent system has become a critical, yet often enigmatic aspect of multi-agent system design~\cite{DBLP:conf/iclr/ChenSZ0YCYLHQQC24,DBLP:journals/corr/abs-2502-08691}.
While intuition and some preliminary studies~\cite{DBLP:journals/corr/abs-2406-07155,DBLP:conf/nips/LiHIKG23} suggest that larger groups would lead to a better reasoning performance, the actual effectiveness of scaling within the context of collaborative expertise specialization and the potential computation-performance trade-off, are not well understood. 
Motivated by the lack of clarity on scaling effects in collaborative expertise specialization, we specifically study how performance scales in multi-agent systems composed of specialized experts. Our systematic experiments involved incrementally increasing the system scale to examine potential scaling laws. 
The results uncover non-linear dynamics; specifically, \textbf{adding more experts tends to improve the collective reasoning ability of the system. 
This positive trend holds regardless of whether the larger system scale contains greater viewpoint diversity or a more comprehensive workflow structure}, indicating a general benefit to increasing the number of expert agents and encouraging such designs for enhanced system performance.
Furthermore, our analysis of the computational trade-offs associated with system scaling reveals that, while the system would benefit from the expansion, there remains a critical need for more efficient communication protocols between agents for more scalable and cost-effective multi-agent reasoning.

\section{Related Works}
\subsection{Multi-Agent Collaboration}
Multi-Agent Collaboration adopts multiple LLMs to solve the problem collaboratively.
Abundant researches have investigated the multi-agent collaboration framework to improve decision-making capability of the system~\cite{DBLP:conf/acl/WangWSTS24,DBLP:conf/emnlp/Liang0JW00Y0T24,DBLP:conf/icml/Du00TM24}. 
In addition to collaboration among LLMs, several researchers instruct the agents to cooperate in a workflow to study the multi-agent systems' ability of solving real world challenges~\cite{DBLP:journals/corr/abs-2405-02957, DBLP:journals/corr/abs-2412-14161,DBLP:journals/corr/abs-2408-08089}.
While \citet{DBLP:journals/corr/abs-2406-07155}, \citet{DBLP:journals/corr/abs-2411-11581} and \citet{DBLP:conf/acl/WangYZQSBN024} has investigate the effect of varying the scale of multi-agent system on reasoning and simulation, prior researches have not systematically examined the interplay between collective expertise specialization, collaboration mechanisms, and the impact of system scale simultaneously.
In this work, we conduct extensive experiments to formally analyze the influence of these three critical dimensions on multi-agent collaborative reasoning. 
Our findings provide actionable insights toward more effective system design.

\begin{figure*}[htp]
    \centering
    \includegraphics[width=\linewidth]{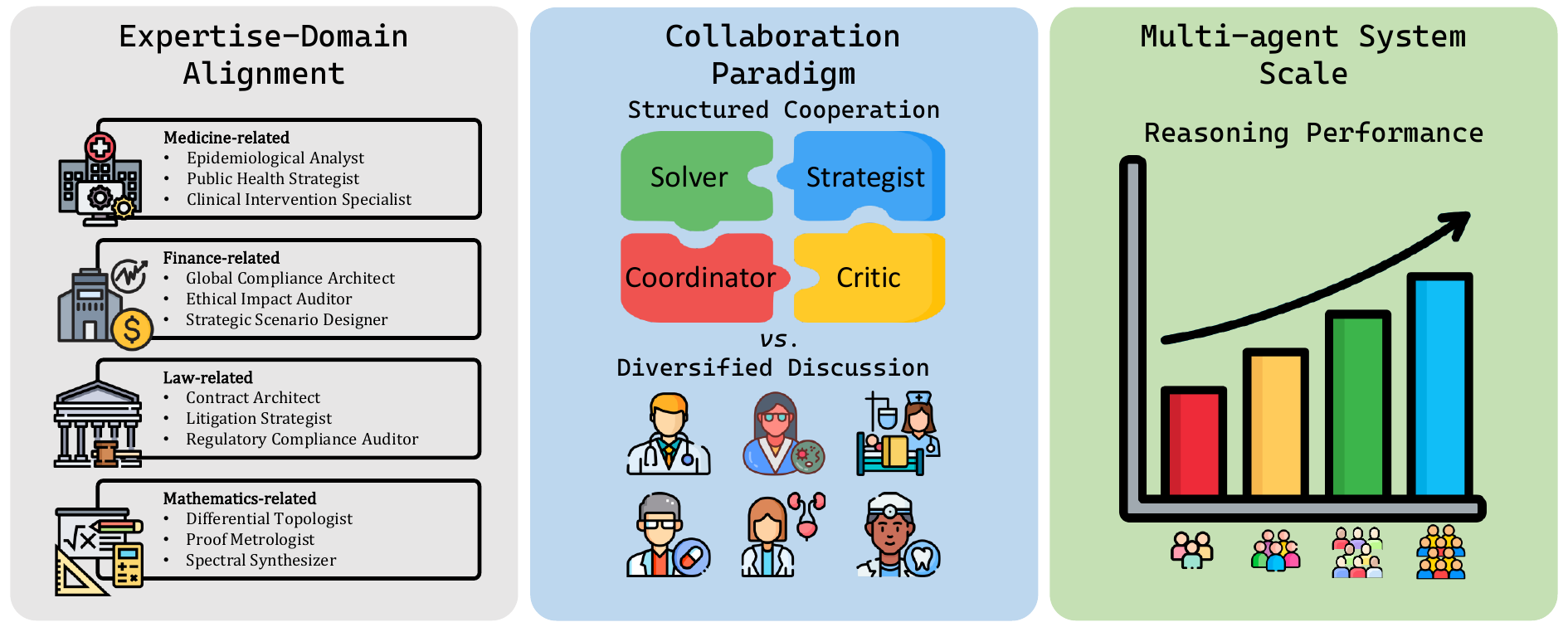}
    \caption{Demonstration of three key dimensions characterizing research on multi-agent collaborative reasoning systems. 
    }
    \label{fig:perspectives}
    \vspace{-0.4cm}
\end{figure*}
\subsection{LLMs as Domain Experts}
The rapid evolution of large language models (LLMs) has endowed them with vast repositories of domain-specific knowledge, enabling their application across a wide range of expert tasks.
Recent research has explored the potential of LLMs to emulate specific personas by conditioning them on detailed character profiles~\cite{DBLP:journals/corr/abs-2406-20094,DBLP:journals/corr/abs-2407-18416,DBLP:journals/corr/abs-2305-14688}. 
These studies demonstrate that by providing LLMs with demographic or role-specific prompts, they can effectively exhibit human-like personality traits and behaviors. 
Furthermore, \citet{DBLP:conf/naacl/KongZCLQSZWD24} and \citet{DBLP:journals/corr/abs-2305-14688} have shown that instructing LLMs to simulate domain experts can enhance their reasoning capabilities in specialized contexts.
Despite these advancements and the growing prominence of multi-agent systems in research, the specific impact of collaborative expertise specialization on reasoning performance remains underexplored. 
In this paper, through meticulously designed experiments, we systematically investigate the impact of expertise specialization within multi-agent reasoning systems. 
Our findings reveal that simulating specialized roles significantly enhances performance on tasks requiring contextual reasoning, while showing limited influence on those primarily dependent on factual recall or mathematical deduction.

\section{Preliminary}
\subsection{Problem Setup}
Formally, given a multi-agent system $\mathcal{M}_n = \{\mathcal{A}_1,\mathcal{A}_2,...,\mathcal{A}_n\}$ where n indicates the number of agents inside the system and $\mathcal{A}_i$ represents the i-th agent of the system, a query $\mathcal{Q}$, and a set of candidate options $\mathcal{S}$.
A multi-agent system reasoning process could be expressed as:
$$\mathcal{Y} = \mathcal{F}(\mathcal{A}_1(\mathcal{Q},\mathcal{S}),\mathcal{A}_2(\mathcal{Q},\mathcal{S}),...,\mathcal{A}_n(\mathcal{Q},\mathcal{S}))$$
where $\mathcal{Y}$ stands for the final answer generated by the system. $\mathcal{A}_i(\mathcal{Q},\mathcal{S})$ represents the answer of agent i, $\mathcal{F}$ stands for the communication protocol manually customized by the design of the system which aggregate the answer of each agents into the final answer.
Typically, it could be majority vote, debate, etc~\cite{DBLP:journals/corr/abs-2502-19130,DBLP:journals/corr/abs-2409-14051}.
In our specific setup, we adopt a sequential processing communication mechanism inspired by~\citet{DBLP:journals/corr/abs-2406-07155} to prevent context explosion~\cite{DBLP:conf/nips/LiuWWXQDDCYQ24,DBLP:conf/iclr/0008PWM0LSBSC24}. 
In this mechanism, for $i=2, ..., n$, agent $\mathcal{A}_i$ receives the complete output generated by the immediately preceding agent $\mathcal{A}_{i-1}$.
In contrast, from the preceding agents \{$\mathcal{A}_1, ..., \mathcal{A}_{i-2}$\}, $\mathcal{A}_i$ receives only the final answers.
The detailed communication algorithm could be found in Appendix~\ref{app:algo} algorithm~\ref{appendix:Communication Algorithm}.

\subsection{Dataset}
\label{Dataset_Preliminary}
For our experiments, we select four distinct domains from MMLU-pro~\cite{DBLP:conf/nips/WangMZNCGRAHJLK24}: Math, Health, Business, and Law. 
These four domains were selected for being representative and frequently studied in contemporary multi-agent reasoning research~\cite{DBLP:journals/corr/abs-2306-16092,DBLP:conf/nips/LeiZZPD24,DBLP:journals/corr/abs-2502-08916}.
We further classify these four domains into three categories based on the primary reasoning type required:
(1) \textbf{Mathematical Reasoning}: Domains requiring formal mathematical deduction to derive the answer.
(2) \textbf{Factual Recall Reasoning}: Domains primarily requiring the recall of domain-specific factual knowledge, seldom needing extensive reasoning steps.
(3) \textbf{Contextual Reasoning}: Domains requiring not only the retrieval of relevant expert knowledge but also its application within the reasoning process of specific scenarios or contexts.
This choice of evaluation domains and fine-grained classification of their reasoning types allow us to investigate the effects of collaborative expertise specialization on multi-agent system  from a more systematic manner.

\subsection{Collaborative Expertise Specialization}
In this paper, we primarily studied the effect of collaborative expertise specialization on better multi-agent system design from the perspective of expert-domain alignment, collaboration paradigms and system scale.
To formalize the role and responsibility of the agents in the multi-agent system, we define each expert to be of the following format:
$$\mathcal{A}_i \leftarrow (\textit{EG},\textit{FR},\textit{R},\textit{ID})$$
Where $\mathcal{A}_i$ stands for the agent i, \textit{EG}, \textit{FR} and \textit{R} represent Expert Group, Formal Role, Responsibility respectively.
ID represents an agent's index within the group of all agents who share the same specific expert role.

\subsection{General Experiment Setup}
As detailed in Section~\ref{Dataset_Preliminary}, we select 4 representative domains from MMLU-pro to investigate the effects of collaborative expertise specialization.
To be consistent with all experiments, we utilize DeepSeek-R1-Distill-Qwen-7B~\cite{DBLP:journals/corr/abs-2501-12948} as the foundational model for all experiments.
Each agent is initialized with its specific expert description and responsibilities via its system prompt, while the task instance is provided through the user prompt.
The detailed prompts could be found in Appendix~\ref{app:system_prompt}.
All experiments adopt accuracy as the evaluation metric.
\section{Leveraging with the ``Right'' Agent}
Expertise specialization is a widely adopted technique in agent research, demonstrably enhancing the reasoning capabilities of Large Language Models (LLMs) within specific domains~\cite{DBLP:journals/corr/abs-2405-02957}. 
While the benefits of specialization for individual agents are well-established, the effect of collaborative expertise specialization on the collective reasoning performance of multi-agent systems remains underexplored. 
This section presents our experimental investigation into this critical area, designed to unveil how different collaborative specialization configurations influence the reasoning capabilities of multi-agent systems.
\label{Sec:RQ1}
\subsection{Setup}
Considering the primary principle of multi-agent reasoning system is to incorporate more diverse agent viewpoints and integrate them in the final answer~\cite{DBLP:conf/emnlp/Liang0JW00Y0T24}, in our experiments, we adopt diversity-driven collaboration paradigm where we distribute each agent with a specific domain expert configuration and instruct them to generate responses based on their expertise.
At this stage, we fix the size of the multi-agent reasoning system to be 3 for controllable computational cost.
We employ GPT-4o~\cite{openai2024gpt4ocard}. 
The model is prompted with the target expert domain, the group size as inputs. 
The detailed prompts utilized for this automated role generation process are provided in Appendix~\ref{app:Expert Generation Prompts}. 


\subsection{``Right'' Expertise Helps Reasoning}
\textbf{Our experiments demonstrate a clear performance advantage when the collaborative expertise specialization of the multi-agent system aligns with the domains of the downstream task.}
Misaligned expertise configurations often underperform compared to aligned ones.
This primary finding is quantitatively supported by the results presented in Table~\ref{Tab:Domain_alignment Result}.
Specifically, in 75\% of the aligned cases (diagonal entries), the system achieved the highest accuracy compared to configurations where the agent group simulates expertise from other domains for the same task.

To gain a more nuanced understanding of when expertise alignment is most beneficial, we analyzed performance according to the primary reasoning type required by each domain, as categorized in Section~\ref{Dataset_Preliminary}.
Our analysis reveals that the benefits of expertise alignment are most pronounced for tasks demanding contextual reasoning. 
Systems operating on Health and Law domains exhibits an average relative performance improvement of 6.75\% when expertise was correctly aligned, compared to the misaligned configurations which perform the second best for those tasks.
Conversely, for domains requiring mathematical reasoning (Math/Business), the specialized experts yield only marginal gains or even degradation relative to misaligned configurations. 
We hypothesize this divergence stems from the inherent strengths of LLMs on math. 
These models often possess robust mathematical reasoning capabilities due to extensive pre-training, potentially reducing the added value of specialized  agents. 
Contextual reasoning tasks, however, appear to benefit more from the structured integration of specialized perspectives provided by the multi-agent reasoning system since applying domain knowledge in these contexts often requires nuanced interpretation, synthesis of information, and reasoning beyond direct mathematical deduction.

{\renewcommand{\arraystretch}{1.5}
\begin{table}[]
\small
\resizebox{\linewidth}{!}{%
\begin{tabular}{@{}l|cccc|cc@{}}
\toprule
\textbf{Dom.$\backslash$Exp.} & Math & Fina & Med & Law & $\Delta_h$ & $\Delta_{abs}$ \\ \midrule
Math     & \textbf{78.0} & 76.3 & 76.3 & \underline{76.4} & 2.1\% & $1.6\uparrow$ \\
Business & \textbf{65.4} & \underline{64.3} & 62.4 & 62.4 & -1.7\% & $1.1 \downarrow$ \\
Health   & \underline{28.9} & 26.8 & \textbf{30.4} & 26.1 & 5.2\% & $1.5 \uparrow$ \\
Law      & 18.3 & \underline{19.2} & 18.5 & \textbf{20.8} & \textbf{8.3}\% & $1.6 \uparrow$ \\ \bottomrule
\end{tabular}
}
\caption{This table shows the performance impact of collaborative expertise specialization for different expert groups across various domains. 
"Dom." and "Exp." abbreviate Domain and Expert Group, respectively. 
$\Delta_{rel}$/$\Delta_{abs}$ indicate the relative/absolute performance improvement of the domain-aligned expert group compared to the best-performing alternative group respectively.}
\label{Tab:Domain_alignment Result}
\end{table}
}


\subsection{Analysis on Expert-Domain Alignment}
\begin{figure}[t]
    \centering
    \includegraphics[width=1\linewidth]{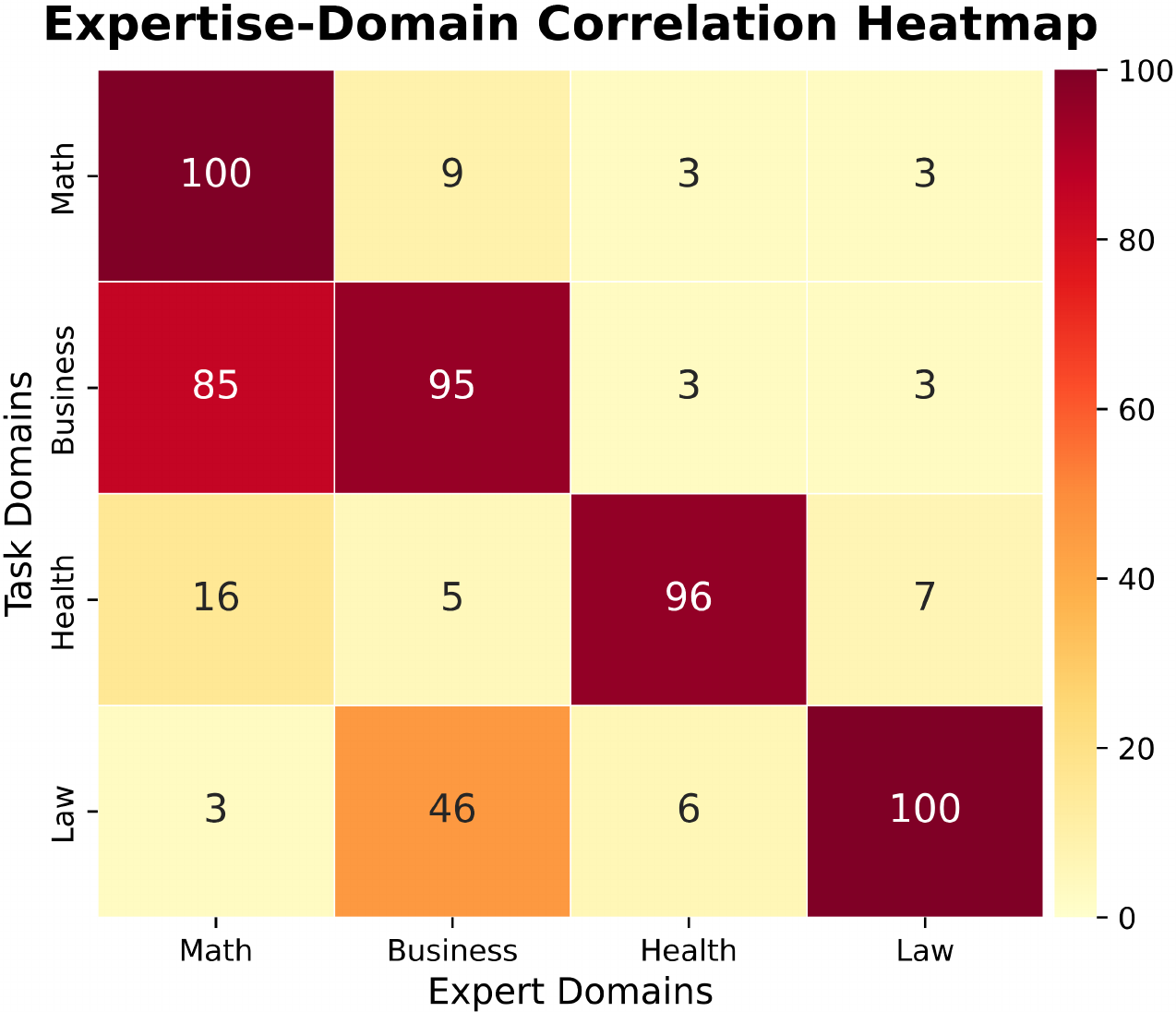}
    \caption{Heatmap illustrating the correlation between specialized group expertise and task domains. Deeper colors indicate stronger correlations.}
    \label{fig:Exp-Dom Cor}
    \vspace{-0.4cm}
\end{figure}

Furthermore, our experimental results reveal a positive correlation between how well the simulated group expertise aligns with (or matches) the downstream task domain and the observed performance gain.
This relationship is visualized in the task-domain relevance analysis presented in Figure~\ref{fig:Exp-Dom Cor}. 
Specifically, configurations where the simulated expertise is more relevant to the target task domain tend to yield greater performance improvements compared to less relevant configurations.

To quantify this expertise-task relevance, we first establish a relevance matrix.
We randomly sample 100 instances from each of the four primary task domains. 
For each instance, we prompt Deepseek-V3~\cite{DBLP:journals/corr/abs-2501-12948} to identify a list of 2-3 key expertise domains pertinent to solving the task. 
We then aggregate these identified candidate domains across all instances within each primary task domain. 
The relevance scores are calculated by counting the occurrences where a specific knowledge domain (e.g., Business) is deemed relevant for tasks in a primary domain (e.g., Math).
These frequencies form a relevance matrix, visualized as a heatmap in Figure~\ref{fig:Exp-Dom Cor}, where deeper color indicate higher relevance scores.

Comparing this relevance heatmap with the performance results, we observe a consistent pattern supporting our initial finding\textemdash Higher expertise-domain relevance, indicated by deeper colors in the heatmap entries, generally corresponds to better reasoning performance. 
Many cells with high relevance scores in Figure~\ref{fig:Exp-Dom Cor} correspond to performance that are bolded or underlined in Table~\ref{Tab:Domain_alignment Result}, signifying the best or second-best performance among group expertise specialization performance for that task domain.
Conversely, low relevance scores typically correspond to misaligned configurations which barely demonstrate distinct advantages conferred by their specific (misaligned) expertise.

Our findings further support the established use of collective expertise specialization in multi-agent reasoning systems, while simultaneously highlighting the critical importance of aligning expertise design with the specific requirements of the target downstream domains, paving a fundamental guidance for future specialization technique application in multi-agent reasoning system design.

\section{Structured Collaboration versus Diversified Discussion}
A further critical consideration in multi-agent system design is the selection of an effective collaboration paradigm. 
Even when individual agents possess appropriate domain knowledge, the overarching mechanism governing their interaction can impact overall system performance. 
In this section, we present comparative experiments designed to analyze these distinct collaboration paradigms. 
Our objective is to investigate their potential advantages, thereby providing empirically grounded insights to guide the design of more effective multi-agent system.
\subsection{Setup}
\begin{figure}[t]
    \centering
    \includegraphics[width=0.95\linewidth]{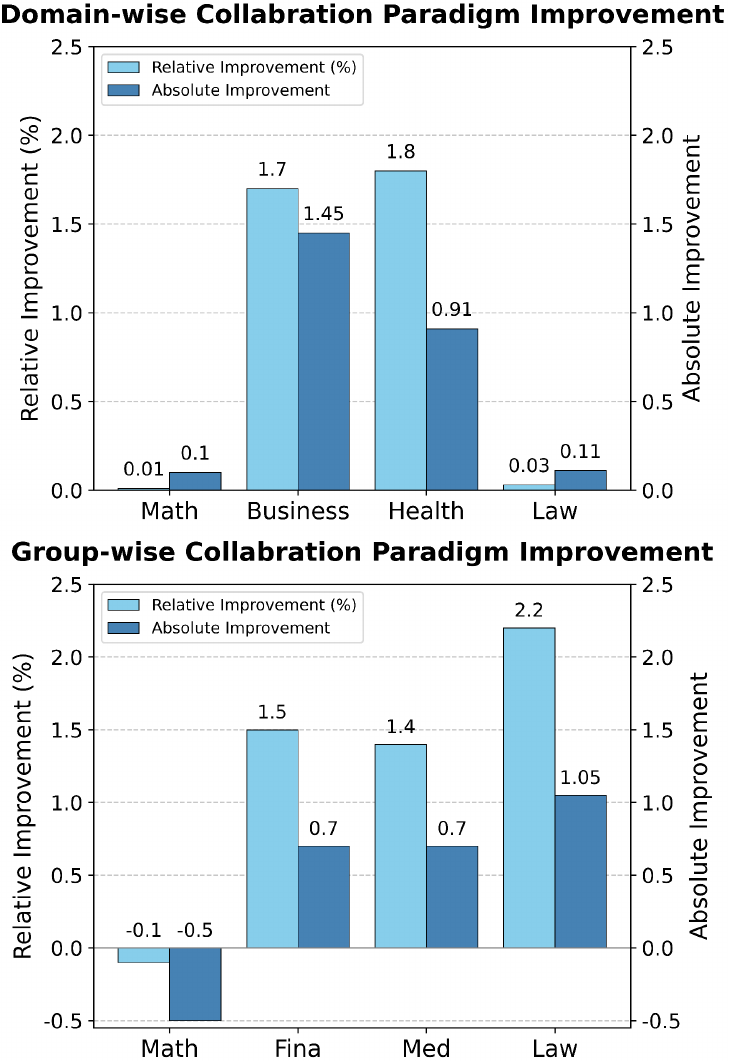}
    \caption{Comparative analysis of diversity-driven versus structured workflow collaboration paradigms.
    Positive values signify Diversity-Driven's advantage over Structured Workflow.}
    \label{fig:RQ2}
    \vspace{-0.4cm}
\end{figure}

Our analysis leverages the results presented in Figure~\ref{fig:RQ2}, where we demonstrate both domain-wise and group-wise comparisons for a comprehensive overview.
The distinction between paradigms are illustrated as follows:\\
\textbf{Diversity-Driven Collaboration:} 
    This paradigm emphasizes assigning agents highly specialized, fine-grained expertise within a broader domain (e.g., specific sub-fields of Laws). 
    The objective is to foster collaboration through the integration of diverse, complementary knowledge perspectives during the reasoning process. 
    Each agent contributes deep expertise from a narrow viewpoint.\\
\textbf{Structured Workflow Collaboration:}
Conversely, this paradigm assigns roles based on distinct functional responsibilities within a predefined problem-solving process, in our case, solver, critic and coordinator.
    Collaboration centers on agents executing specific steps and refining intermediate outputs based on their functional role, rather than primarily contributing unique domain knowledge specializations.
    The differentiation between agents stems from their function within the workflow.

To ensure a plausible, accurate generation of expert role descriptions, we continue to employ GPT-4o with collaboration paradigm as extra input.  


\subsection{Diversity Matters in Discussion}
Our primary finding is that the diversity-driven paradigm generally yields superior performance compared to the structured workflow paradigm. 
This advantage holds true both when considering performance from both domain-wise and group-wise perspectives.

A domain-wise analysis, depicted in Figure~\ref{fig:RQ2}, confirms this trend. 
Irrespective of the domain's primary reasoning type categorized in section~\ref{Dataset_Preliminary}, the diversity-driven approach consistently results in performance gains over structured workflow. 
Notably, the most substantial improvements are observed in business and health domains, which demonstrate an average relative performance increase of 1.75\% under diversity-driven paradigm.
This indicates the potential of expertise with finer-granularity perform well across different domains.

\begin{figure}[t]
    \centering
    \includegraphics[width=\linewidth]{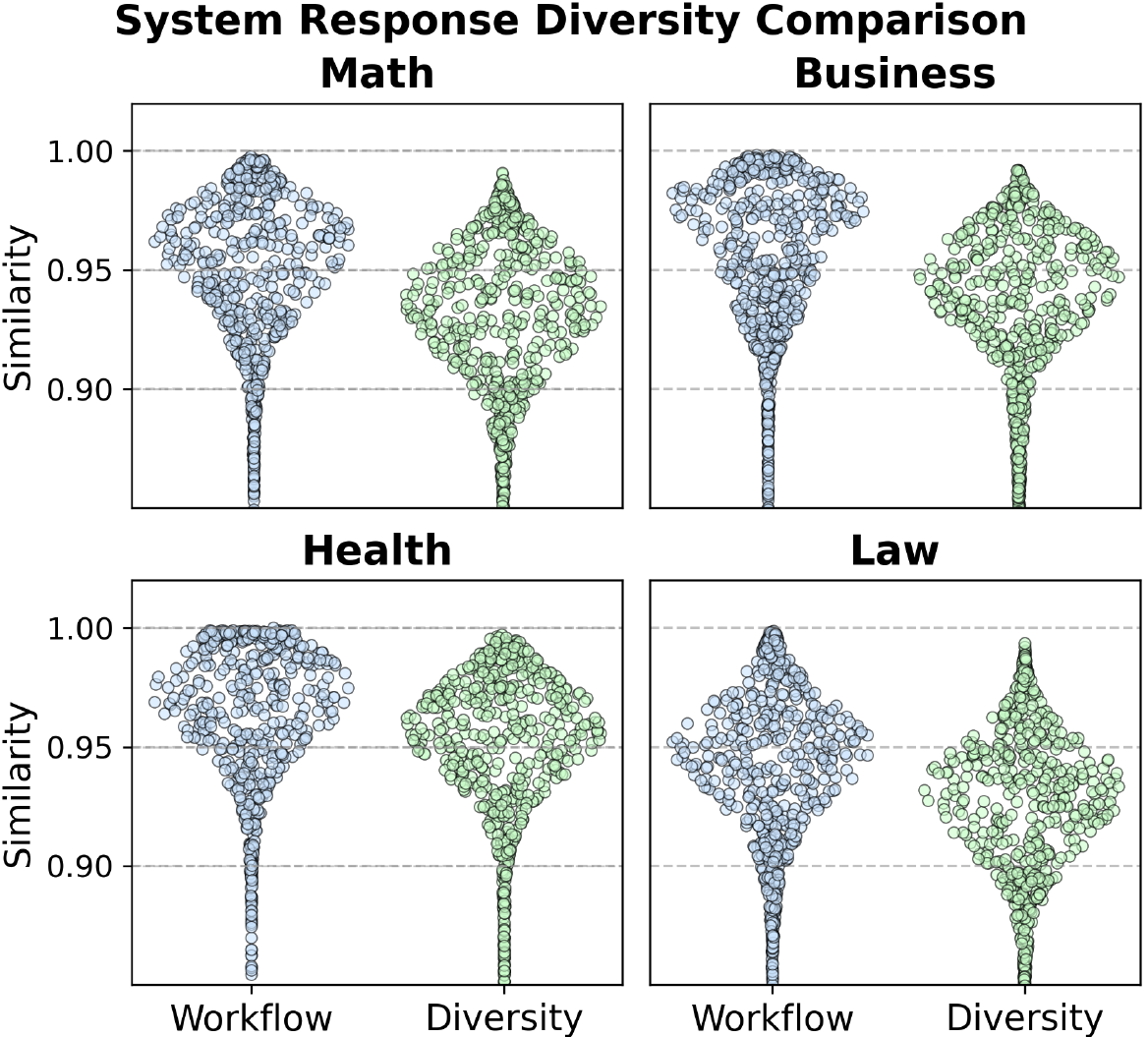}
    \caption{Illustration of response diversity across four distinct domains, where lower inter-agent response similarity corresponds to higher diversity.}
    \label{fig:diversity-similarity}
    \vspace{-0.4cm}
\end{figure}

Examining the results from group-wise perspective further supports this conclusion. 
With the exception of math expert group, all other specialized groups achieve higher average performance across all task domains when employing diversity-driven paradigm. 
When including the math group, the overall average relative performance improvement facilitated by the diversity-driven approach across all groups was 1.25\%, indicating consistent benefits regardless of the task domain encountered.

Synthesizing these observations, the diversity-driven collaboration paradigm demonstrates a consistent performance advantage over structured workflow collaboration paradigm across both different tested domains and distinct expertise configurations.
This suggests that multi-agent systems could benefit significantly from collaboration structures that emphasize fine-grained expertise allocation which stimulates viewpoint diversity, providing an empirical basis for future research directions in designing multi-agent reasoning system's collaboration pattern.

\subsection{Analysis on Response Diversity}
To quantitatively characterize how the collaboration paradigm influences the diversity of agent contributions, we further design a response diversity analysis. 
We leverage semantic embeddings derived from Sentence-BERT~\cite{DBLP:conf/emnlp/ReimersG19}. 
For each task instance solved by the multi-agent system, we generate embeddings for the output of each agent.
We then measure the internal diversity of the system's responses by calculating the pairwise cosine similarity between the embeddings of outputs from different agents. 
This provides a measure of how semantically distinct the contributions are at different stages. 

The distributions of these pairwise similarity scores for both the diversity-driven and structured workflow paradigms are presented in Figure~\ref{fig:diversity-similarity}. 
The results clearly indicate that, the pairwise cosine similarity values are consistently lower for the diversity-driven collaboration paradigm compared to the structured workflow paradigm.
This finding demonstrates that the diversity-driven approach, which emphasizes fine-grained expertise, fosters greater semantic diversity among agent responses throughout the collaborative reasoning, further confirming that enhancing the diversity of perspectives within a multi-agent system would be a key factor in improving its overall reasoning performance.

\begin{figure}[t]
    \centering
    \includegraphics[width=0.95\linewidth]{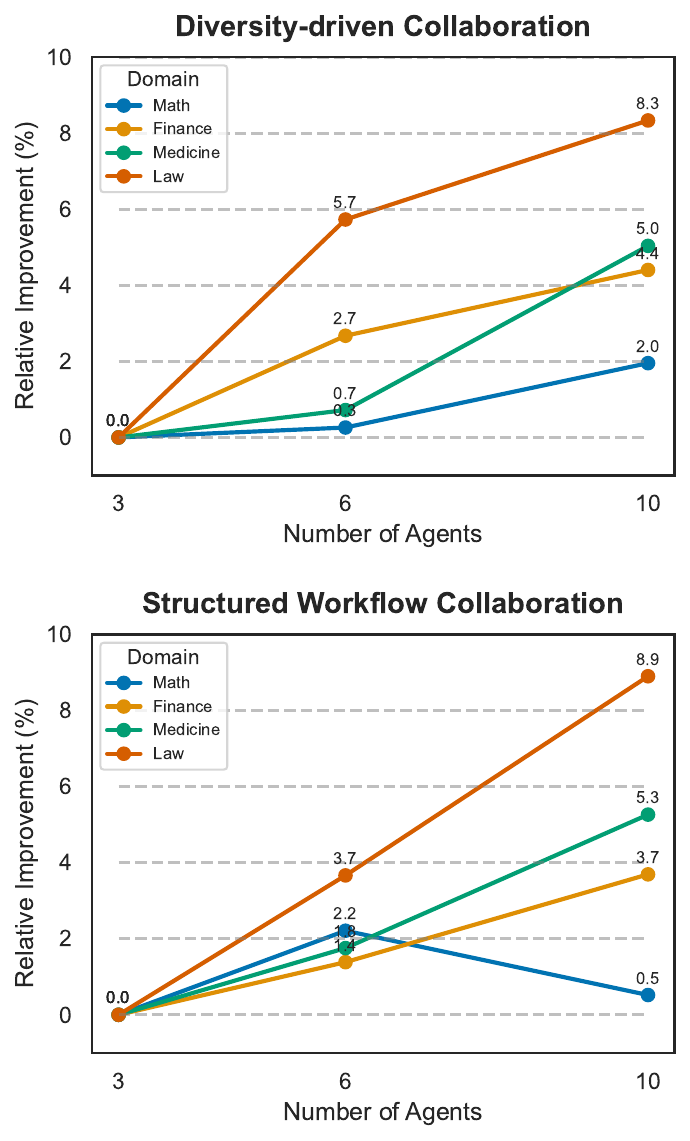}
    \caption{Domain-wise relative performance improvement resulting from scaling up the multi-agent system (3,6,10), shown for two different cooperation mechanisms.}
    \label{fig:scale_improvement}
    \vspace{-0.4cm}
\end{figure}


\section{Scaling Up Reasoning Experts}
Finally, the most complicated dimension in designing multi-agent systems to foster collective intelligence is the system scale. 
While the deployment of large-scale multi-agent systems for simulating social behaviors has received considerable attention, the implications of scaling under collaborative expertise specialization setup remain unexplored. 
This section details our investigation into the effects of varying system scale on both the reasoning performance of multi-agent systems and the associated computational trade-offs. 
We aim to elucidate how increasing the number of agents influences collective reasoning efficacy and to call for a better communication protocol design through our trade-off analysis.
\subsection{Setup}
We expand our experimental setup from 3 agent to systems comprising 6 and 10 agents. 
For these larger system sizes, we systematically replicate the experiments previously introduced, allowing for a direct comparison across different scales.

Generating coherent and appropriately specialized expert role configurations for these larger systems requires extending the initial configurations of the 3 agent system and we continue to leverage GPT-4o for this purpose. 
The detailed prompts employed for this role augmentation process are provided in Appendix~\ref{app:Expert Generation Prompts}

\subsection{More Experts, More Intelligent System}
We evaluate the effect of system scale on reasoning performance by comparing the results from larger agent systems against the baseline 3 agent system. 
Specifically, we calculate the domain-wise relative performance difference for the system size of 6 and 10 with respect to system of size 3. 
These relative performance differences are illustrated in Figure~\ref{fig:scale_improvement}.

Our findings reveal a consistent trend: increasing the number of agents generally enhances the multi-agent system's reasoning performance across the evaluated domains, regardless of whether diversity-driven or structured workflow paradigm was employed.
However, the magnitude of this improvement varies significantly by domain.
Corroborating our earlier observations regarding domain-specific analysis in section~\ref{Sec:RQ1}, the performance gains within math domain were marginal, even when scaling up to 10 agents. 
Conversely, domains that necessitate substantial contextual reasoning and knowledge application demonstrate significantly larger performance improvements with increased system scale.
This disparity suggests that the benefits derived from incorporating additional agents are most pronounced for tasks requiring the integration of diverse knowledge perspectives or complex, case-specific analysis inherent in non-mathematical reasoning.
For domains characterized by intense mathematical reasoning, simply increasing the number of agents could barely yield diminishing returns.
We believe our finding offers valuable insight for constructing large-scale multi-agent systems intended for diverse domains.


\subsection{Token-Performance Trade-off}

\begin{figure}[t]
    \centering
    \includegraphics[width=\linewidth]{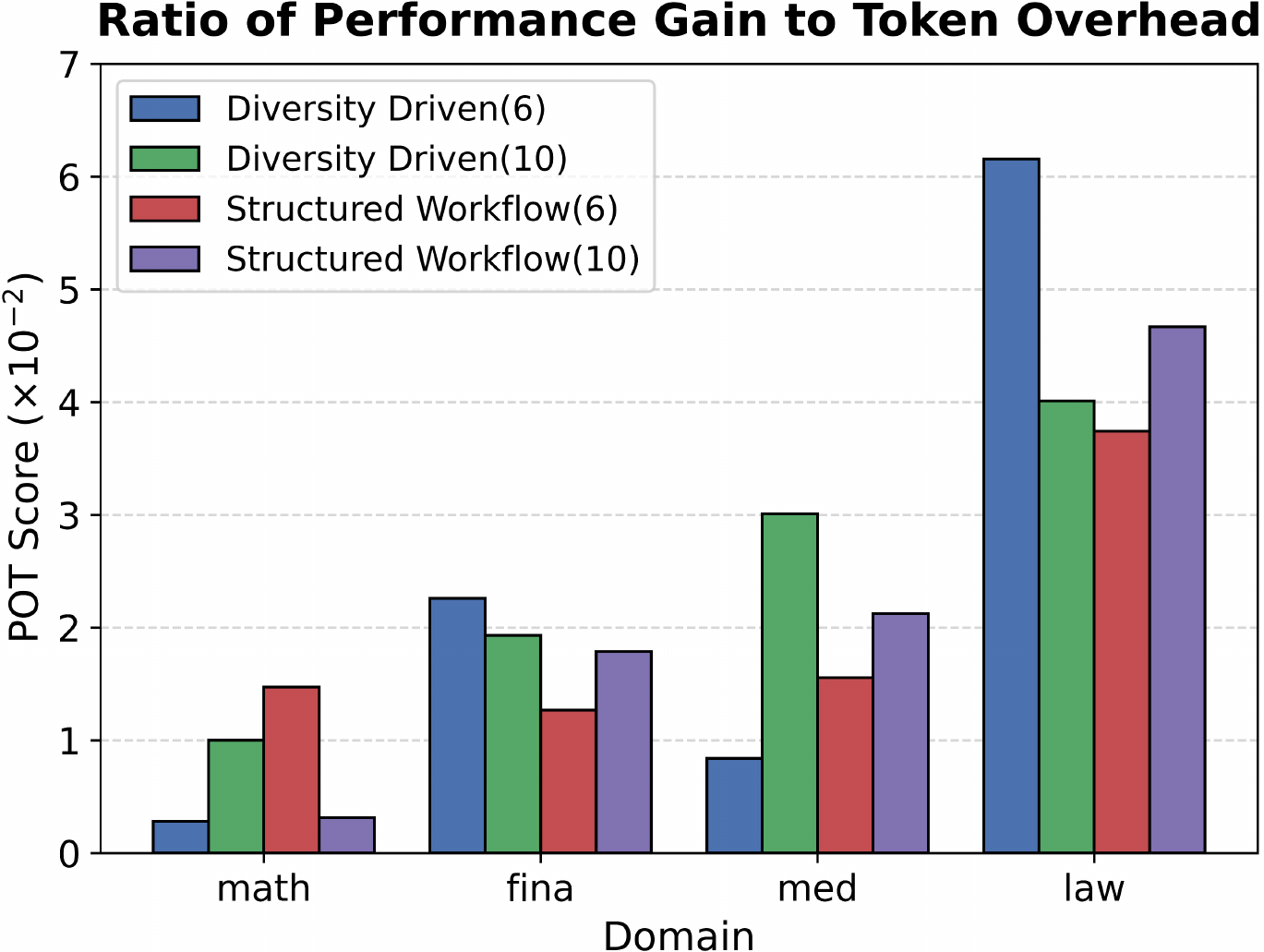}
    \caption{Performance improvement versus token overhead ratio across different domains.
    Both performance and token overhead are measured as relative increases compared to the system of size 3.}
    \label{fig:token_improve}
    \vspace{-0.4cm}
\end{figure}

We further explore the token-performance trade-off inherent in scaling multi-agent reasoning systems by calculating the ratio of performance improvement over token overhead (PoT) with qualitative results presented in Figure~\ref{fig:token_improve}. 
We use the sum of reasoning token and answer token for the calculation of token overhead and all the performance improvement and token consumption overhead are counted against system of size 3.

Our analysis reveals distinct trends both across and within domains. 
Cross-domain comparisons demonstrate that tasks requiring substantial contextual reasoning, such as those in health and law, yield higher PoT ratios. 
This suggests that increasing agent collaboration is particularly beneficial in these areas, as greater token consumption during the reasoning process leads to higher performance improvements.
Conversely, mathematical reasoning tasks exhibit only marginal performance gains with additional agents, which implies smaller ensembles can achieve comparable performance with lower computational overhead, making large-scale multi-agent systems unnecessary for these tasks.

For intra-domain analysis, while structured workflows improved PoT in 75\% of domains and diversity-driven approaches in 50\% respectively, the critical finding is that neither collaboration paradigm guarantees a enhanced PoT across all domains tested.
This widespread inconsistency in scaling behavior, regardless of the collaboration paradigm, highlights the pressing need for advancements in multi-agent communication protocols to achieve more stable and predictable performance enhancements as system complexity increases.

\section{Conclusions}
In conclusion, this paper systematically investigate the impact of multi-agent system expertise specialization on collective reasoning intelligence across three key dimensions: expertise-domain alignment, collaboration paradigm, and system scale.
Our empirical results demonstrate that the effectiveness of aligning agent expertise with task domains is highly contingent on the reasoning type required. 
Furthermore, we find that collaboration paradigms emphasizing the integration of diverse perspectives outperform structured task decomposition, with response diversity playing a crucial role. 
Finally, a scaling law behind collaborative expertise specialization in multi-agent system is observed where stacking more experts could result in better reasoning performance.
These findings provide actionable insights for designing specialized multi-agent reasoning systems in future researches and underscore the critical need for developing more efficient coordination protocol as systems scale.

\section*{Limitations}
The limitations of the paper concern following several dimensions.
1. Since we focus on the reasoning capability of multi-agent system, we adopt MMLU-pro as our evaluation dataset which consist of massive reasoning tasks instead of tasks like coding.
2. To be closer to real-world scenario, we utilize only four domains (Math, Business, Health and Law) which are most focused by mainstream researches while leaving other domains underexplored.
3. To simplify the research setup and to yield more stable conclusion, we conduct our experiments on only one message propagation mechanism while overlook the potential impact could be brought by other communication mechanisms.
4. For controllable computational cost, we select DeepSeek-R1-Distilled-Qwen-7B as our base model for all the experiments, leaving multi-agent system design for larger model for future studies.

\section*{Ethics Statement}
Our study involves publicly available datasets and use Large Language Models through APIs. 
Consequently, the ethical considerations of this paper could be listed as follow:\\
\textbf{Datasets:} We use publicly available datasets only for academic research purpose. 
We guarantee no personal data has been involved.\\
\textbf{LLMs API: } Our application of LLMs conform API provider's policy strictly, maintaining fair use and respecting intellectual property.\\
\textbf{Transparency: } We provide detailed descriptions of our method and the prompts used in our experiments, in line with standard practices in the research community. 
We will also make our code publicly available upon acceptance.


\bibliography{custom}

\begin{thebibliography}{38}
\providecommand{\natexlab}[1]{#1}

\bibitem[{Chan et~al.(2024)Chan, Wang, Yu, Mi, and Yu}]{DBLP:journals/corr/abs-2406-20094}
Xin Chan, Xiaoyang Wang, Dian Yu, Haitao Mi, and Dong Yu. 2024.
\newblock \href {https://doi.org/10.48550/ARXIV.2406.20094} {Scaling synthetic data creation with 1,000,000,000 personas}.
\newblock \emph{CoRR}, abs/2406.20094.

\bibitem[{Chen et~al.(2024{\natexlab{a}})Chen, Fan, Gong, Xie, Li, Liu, Li, Qu, Ni, and Yang}]{DBLP:journals/corr/abs-2408-08089}
Guhong Chen, Liyang Fan, Zihan Gong, Nan Xie, Zixuan Li, Ziqiang Liu, Chengming Li, Qiang Qu, Shiwen Ni, and Min Yang. 2024{\natexlab{a}}.
\newblock \href {https://doi.org/10.48550/ARXIV.2408.08089} {Agentcourt: Simulating court with adversarial evolvable lawyer agents}.
\newblock \emph{CoRR}, abs/2408.08089.

\bibitem[{Chen et~al.(2024{\natexlab{b}})Chen, Saha, and Bansal}]{DBLP:conf/acl/ChenSB24}
Justin~Chih{-}Yao Chen, Swarnadeep Saha, and Mohit Bansal. 2024{\natexlab{b}}.
\newblock \href {https://doi.org/10.18653/V1/2024.ACL-LONG.381} {Reconcile: Round-table conference improves reasoning via consensus among diverse llms}.
\newblock In \emph{Proceedings of the 62nd Annual Meeting of the Association for Computational Linguistics (Volume 1: Long Papers), {ACL} 2024, Bangkok, Thailand, August 11-16, 2024}, pages 7066--7085. Association for Computational Linguistics.

\bibitem[{Chen et~al.(2024{\natexlab{c}})Chen, Su, Zuo, Yang, Yuan, Chan, Yu, Lu, Hung, Qian, Qin, Cong, Xie, Liu, Sun, and Zhou}]{DBLP:conf/iclr/ChenSZ0YCYLHQQC24}
Weize Chen, Yusheng Su, Jingwei Zuo, Cheng Yang, Chenfei Yuan, Chi{-}Min Chan, Heyang Yu, Yaxi Lu, Yi{-}Hsin Hung, Chen Qian, Yujia Qin, Xin Cong, Ruobing Xie, Zhiyuan Liu, Maosong Sun, and Jie Zhou. 2024{\natexlab{c}}.
\newblock \href {https://openreview.net/forum?id=EHg5GDnyq1} {Agentverse: Facilitating multi-agent collaboration and exploring emergent behaviors}.
\newblock In \emph{The Twelfth International Conference on Learning Representations, {ICLR} 2024, Vienna, Austria, May 7-11, 2024}. OpenReview.net.

\bibitem[{Cui et~al.(2023)Cui, Li, Yan, Chen, and Yuan}]{DBLP:journals/corr/abs-2306-16092}
Jiaxi Cui, Zongjian Li, Yang Yan, Bohua Chen, and Li~Yuan. 2023.
\newblock \href {https://doi.org/10.48550/ARXIV.2306.16092} {Chatlaw: Open-source legal large language model with integrated external knowledge bases}.
\newblock \emph{CoRR}, abs/2306.16092.

\bibitem[{DeepSeek{-}AI et~al.(2025)DeepSeek{-}AI, Guo, Yang, Zhang, Song, Zhang, Xu, Zhu, Ma, Wang, Bi, Zhang, Yu, Wu, Wu, Gou, Shao, Li, Gao, Liu, Xue, Wang, Wu, Feng, Lu, Zhao, Deng, Zhang, Ruan, Dai, Chen, Ji, Li, Lin, Dai, Luo, Hao, Chen, Li, Zhang, Bao, Xu, Wang, Ding, Xin, Gao, Qu, Li, Guo, Li, Wang, Chen, Yuan, Qiu, Li, Cai, Ni, Liang, Chen, Dong, Hu, Gao, Guan, Huang, Yu, Wang, Zhang, Zhao, Wang, Zhang, Xu, Xia, Zhang, Zhang, Tang, Li, Wang, Li, Tian, Huang, Zhang, Wang, Chen, Du, Ge, Zhang, Pan, Wang, Chen, Jin, Chen, Lu, Zhou, Chen, Ye, Wang, Yu, Zhou, Pan, and Li}]{DBLP:journals/corr/abs-2501-12948}
DeepSeek{-}AI, Daya Guo, Dejian Yang, Haowei Zhang, Junxiao Song, Ruoyu Zhang, Runxin Xu, Qihao Zhu, Shirong Ma, Peiyi Wang, Xiao Bi, Xiaokang Zhang, Xingkai Yu, Yu~Wu, Z.~F. Wu, Zhibin Gou, Zhihong Shao, Zhuoshu Li, Ziyi Gao, Aixin Liu, Bing Xue, Bingxuan Wang, Bochao Wu, Bei Feng, Chengda Lu, Chenggang Zhao, Chengqi Deng, Chenyu Zhang, Chong Ruan, Damai Dai, Deli Chen, Dongjie Ji, Erhang Li, Fangyun Lin, Fucong Dai, Fuli Luo, Guangbo Hao, Guanting Chen, Guowei Li, H.~Zhang, Han Bao, Hanwei Xu, Haocheng Wang, Honghui Ding, Huajian Xin, Huazuo Gao, Hui Qu, Hui Li, Jianzhong Guo, Jiashi Li, Jiawei Wang, Jingchang Chen, Jingyang Yuan, Junjie Qiu, Junlong Li, J.~L. Cai, Jiaqi Ni, Jian Liang, Jin Chen, Kai Dong, Kai Hu, Kaige Gao, Kang Guan, Kexin Huang, Kuai Yu, Lean Wang, Lecong Zhang, Liang Zhao, Litong Wang, Liyue Zhang, Lei Xu, Leyi Xia, Mingchuan Zhang, Minghua Zhang, Minghui Tang, Meng Li, Miaojun Wang, Mingming Li, Ning Tian, Panpan Huang, Peng Zhang, Qiancheng Wang, Qinyu Chen, Qiushi Du, Ruiqi Ge,
  Ruisong Zhang, Ruizhe Pan, Runji Wang, R.~J. Chen, R.~L. Jin, Ruyi Chen, Shanghao Lu, Shangyan Zhou, Shanhuang Chen, Shengfeng Ye, Shiyu Wang, Shuiping Yu, Shunfeng Zhou, Shuting Pan, and S.~S. Li. 2025.
\newblock \href {https://doi.org/10.48550/ARXIV.2501.12948} {Deepseek-r1: Incentivizing reasoning capability in llms via reinforcement learning}.
\newblock \emph{CoRR}, abs/2501.12948.

\bibitem[{Du et~al.(2024)Du, Li, Torralba, Tenenbaum, and Mordatch}]{DBLP:conf/icml/Du00TM24}
Yilun Du, Shuang Li, Antonio Torralba, Joshua~B. Tenenbaum, and Igor Mordatch. 2024.
\newblock \href {https://openreview.net/forum?id=zj7YuTE4t8} {Improving factuality and reasoning in language models through multiagent debate}.
\newblock In \emph{Forty-first International Conference on Machine Learning, {ICML} 2024, Vienna, Austria, July 21-27, 2024}. OpenReview.net.

\bibitem[{Ghezloo et~al.(2025)Ghezloo, Seyfioglu, Soraki, Ikezogwo, Li, Vivekanandan, Elmore, Krishna, and Shapiro}]{DBLP:journals/corr/abs-2502-08916}
Fatemeh Ghezloo, Mehmet~Saygin Seyfioglu, Rustin Soraki, Wisdom~Oluchi Ikezogwo, Beibin Li, Tejoram Vivekanandan, Joann~G. Elmore, Ranjay Krishna, and Linda~G. Shapiro. 2025.
\newblock \href {https://doi.org/10.48550/ARXIV.2502.08916} {Pathfinder: {A} multi-modal multi-agent system for medical diagnostic decision-making applied to histopathology}.
\newblock \emph{CoRR}, abs/2502.08916.

\bibitem[{Hong et~al.(2024)Hong, Zhuge, Chen, Zheng, Cheng, Wang, Zhang, Wang, Yau, Lin, Zhou, Ran, Xiao, Wu, and Schmidhuber}]{DBLP:conf/iclr/HongZCZCWZWYLZR24}
Sirui Hong, Mingchen Zhuge, Jonathan Chen, Xiawu Zheng, Yuheng Cheng, Jinlin Wang, Ceyao Zhang, Zili Wang, Steven Ka~Shing Yau, Zijuan Lin, Liyang Zhou, Chenyu Ran, Lingfeng Xiao, Chenglin Wu, and J{\"{u}}rgen Schmidhuber. 2024.
\newblock \href {https://openreview.net/forum?id=VtmBAGCN7o} {Metagpt: Meta programming for {A} multi-agent collaborative framework}.
\newblock In \emph{The Twelfth International Conference on Learning Representations, {ICLR} 2024, Vienna, Austria, May 7-11, 2024}. OpenReview.net.

\bibitem[{Hu et~al.(2025)Hu, Chan, Li, and Yin}]{DBLP:conf/coling/HuCL025}
Zhe Hu, Hou~Pong Chan, Jing Li, and Yu~Yin. 2025.
\newblock \href {https://aclanthology.org/2025.coling-main.314/} {Debate-to-write: {A} persona-driven multi-agent framework for diverse argument generation}.
\newblock In \emph{Proceedings of the 31st International Conference on Computational Linguistics, {COLING} 2025, Abu Dhabi, UAE, January 19-24, 2025}, pages 4689--4703. Association for Computational Linguistics.

\bibitem[{Jaech et~al.(2024)Jaech, Kalai, Lerer, Richardson, El{-}Kishky, Low, Helyar, Madry, Beutel, Carney, Iftimie, Karpenko, Passos, Neitz, Prokofiev, Wei, Tam, Bennett, Kumar, Saraiva, Vallone, Duberstein, Kondrich, Mishchenko, Applebaum, Jiang, Nair, Zoph, Ghorbani, Rossen, Sokolowsky, Barak, McGrew, Minaiev, Hao, Baker, Houghton, McKinzie, Eastman, Lugaresi, Bassin, Hudson, Li, de~Bourcy, Voss, Shen, Zhang, Koch, Orsinger, Hesse, Fischer, Chan, Roberts, Kappler, Levy, Selsam, Dohan, Farhi, Mely, Robinson, Tsipras, Li, Oprica, Freeman, Zhang, Wong, Proehl, Cheung, Mitchell, Wallace, Ritter, Mays, Wang, Such, Raso, Leoni, Tsimpourlas, Song, von Lohmann, Sulit, Salmon, Parascandolo, Chabot, Zhao, Brockman, Leclerc, Salman, Bao, Sheng, Andrin, Bagherinezhad, Ren, Lightman, Chung, Kivlichan, O'Connell, Osband, Gilaberte, and Akkaya}]{DBLP:journals/corr/abs-2412-16720}
Aaron Jaech, Adam Kalai, Adam Lerer, Adam Richardson, Ahmed El{-}Kishky, Aiden Low, Alec Helyar, Aleksander Madry, Alex Beutel, Alex Carney, Alex Iftimie, Alex Karpenko, Alex~Tachard Passos, Alexander Neitz, Alexander Prokofiev, Alexander Wei, Allison Tam, Ally Bennett, Ananya Kumar, Andre Saraiva, Andrea Vallone, Andrew Duberstein, Andrew Kondrich, Andrey Mishchenko, Andy Applebaum, Angela Jiang, Ashvin Nair, Barret Zoph, Behrooz Ghorbani, Ben Rossen, Benjamin Sokolowsky, Boaz Barak, Bob McGrew, Borys Minaiev, Botao Hao, Bowen Baker, Brandon Houghton, Brandon McKinzie, Brydon Eastman, Camillo Lugaresi, Cary Bassin, Cary Hudson, Chak~Ming Li, Charles de~Bourcy, Chelsea Voss, Chen Shen, Chong Zhang, Chris Koch, Chris Orsinger, Christopher Hesse, Claudia Fischer, Clive Chan, Dan Roberts, Daniel Kappler, Daniel Levy, Daniel Selsam, David Dohan, David Farhi, David Mely, David Robinson, Dimitris Tsipras, Doug Li, Dragos Oprica, Eben Freeman, Eddie Zhang, Edmund Wong, Elizabeth Proehl, Enoch Cheung, Eric Mitchell,
  Eric Wallace, Erik Ritter, Evan Mays, Fan Wang, Felipe~Petroski Such, Filippo Raso, Florencia Leoni, Foivos Tsimpourlas, Francis Song, Fred von Lohmann, Freddie Sulit, Geoff Salmon, Giambattista Parascandolo, Gildas Chabot, Grace Zhao, Greg Brockman, Guillaume Leclerc, Hadi Salman, Haiming Bao, Hao Sheng, Hart Andrin, Hessam Bagherinezhad, Hongyu Ren, Hunter Lightman, Hyung~Won Chung, Ian Kivlichan, Ian O'Connell, Ian Osband, Ignasi~Clavera Gilaberte, and Ilge Akkaya. 2024.
\newblock \href {https://doi.org/10.48550/ARXIV.2412.16720} {Openai o1 system card}.
\newblock \emph{CoRR}, abs/2412.16720.

\bibitem[{Kaesberg et~al.(2025)Kaesberg, Becker, Wahle, Ruas, and Gipp}]{DBLP:journals/corr/abs-2502-19130}
Lars~Benedikt Kaesberg, Jonas Becker, Jan~Philip Wahle, Terry Ruas, and Bela Gipp. 2025.
\newblock \href {https://doi.org/10.48550/ARXIV.2502.19130} {Voting or consensus? decision-making in multi-agent debate}.
\newblock \emph{CoRR}, abs/2502.19130.

\bibitem[{Kong et~al.(2024)Kong, Zhao, Chen, Li, Qin, Sun, Zhou, Wang, and Dong}]{DBLP:conf/naacl/KongZCLQSZWD24}
Aobo Kong, Shiwan Zhao, Hao Chen, Qicheng Li, Yong Qin, Ruiqi Sun, Xin Zhou, Enzhi Wang, and Xiaohang Dong. 2024.
\newblock \href {https://doi.org/10.18653/V1/2024.NAACL-LONG.228} {Better zero-shot reasoning with role-play prompting}.
\newblock In \emph{Proceedings of the 2024 Conference of the North American Chapter of the Association for Computational Linguistics: Human Language Technologies (Volume 1: Long Papers), {NAACL} 2024, Mexico City, Mexico, June 16-21, 2024}, pages 4099--4113. Association for Computational Linguistics.

\bibitem[{Lei et~al.(2024)Lei, Zhang, Zuo, Payani, and Ding}]{DBLP:conf/nips/LeiZZPD24}
Bin Lei, Yi~Zhang, Shan Zuo, Ali Payani, and Caiwen Ding. 2024.
\newblock \href {http://papers.nips.cc/paper\_files/paper/2024/hash/5fcedec09977357f32e8e0ec8957073b-Abstract-Conference.html} {{MACM:} utilizing a multi-agent system for condition mining in solving complex mathematical problems}.
\newblock In \emph{Advances in Neural Information Processing Systems 38: Annual Conference on Neural Information Processing Systems 2024, NeurIPS 2024, Vancouver, BC, Canada, December 10 - 15, 2024}.

\bibitem[{Li et~al.(2024{\natexlab{a}})Li, Tan, Qian, Li, Chaudhary, Hu, and Shen}]{DBLP:journals/corr/abs-2411-03284}
Dawei Li, Zhen Tan, Peijia Qian, Yifan Li, Kumar~Satvik Chaudhary, Lijie Hu, and Jiayi Shen. 2024{\natexlab{a}}.
\newblock \href {https://doi.org/10.48550/ARXIV.2411.03284} {Smoa: Improving multi-agent large language models with sparse mixture-of-agents}.
\newblock \emph{CoRR}, abs/2411.03284.

\bibitem[{Li et~al.(2023)Li, Hammoud, Itani, Khizbullin, and Ghanem}]{DBLP:conf/nips/LiHIKG23}
Guohao Li, Hasan Hammoud, Hani Itani, Dmitrii Khizbullin, and Bernard Ghanem. 2023.
\newblock \href {http://papers.nips.cc/paper\_files/paper/2023/hash/a3621ee907def47c1b952ade25c67698-Abstract-Conference.html} {{CAMEL:} communicative agents for "mind" exploration of large language model society}.
\newblock In \emph{Advances in Neural Information Processing Systems 36: Annual Conference on Neural Information Processing Systems 2023, NeurIPS 2023, New Orleans, LA, USA, December 10 - 16, 2023}.

\bibitem[{Li et~al.(2024{\natexlab{b}})Li, Wang, Zhang, Li, Lai, Kang, Ma, and Liu}]{DBLP:journals/corr/abs-2405-02957}
Junkai Li, Siyu Wang, Meng Zhang, Weitao Li, Yunghwei Lai, Xinhui Kang, Weizhi Ma, and Yang Liu. 2024{\natexlab{b}}.
\newblock \href {https://doi.org/10.48550/ARXIV.2405.02957} {Agent hospital: {A} simulacrum of hospital with evolvable medical agents}.
\newblock \emph{CoRR}, abs/2405.02957.

\bibitem[{Li et~al.(2024{\natexlab{c}})Li, Du, Zhang, Hou, Grabowski, Li, and Ie}]{DBLP:conf/emnlp/LiDZHGLI24}
Yunxuan Li, Yibing Du, Jiageng Zhang, Le~Hou, Peter Grabowski, Yeqing Li, and Eugene Ie. 2024{\natexlab{c}}.
\newblock \href {https://aclanthology.org/2024.findings-emnlp.427} {Improving multi-agent debate with sparse communication topology}.
\newblock In \emph{Findings of the Association for Computational Linguistics: {EMNLP} 2024, Miami, Florida, USA, November 12-16, 2024}, pages 7281--7294. Association for Computational Linguistics.

\bibitem[{Liang et~al.(2024)Liang, He, Jiao, Wang, Wang, Wang, Yang, Shi, and Tu}]{DBLP:conf/emnlp/Liang0JW00Y0T24}
Tian Liang, Zhiwei He, Wenxiang Jiao, Xing Wang, Yan Wang, Rui Wang, Yujiu Yang, Shuming Shi, and Zhaopeng Tu. 2024.
\newblock \href {https://aclanthology.org/2024.emnlp-main.992} {Encouraging divergent thinking in large language models through multi-agent debate}.
\newblock In \emph{Proceedings of the 2024 Conference on Empirical Methods in Natural Language Processing, {EMNLP} 2024, Miami, FL, USA, November 12-16, 2024}, pages 17889--17904. Association for Computational Linguistics.

\bibitem[{Liu et~al.(2024{\natexlab{a}})Liu, Wang, Huang, Xu, Zeng, Jiang, Yang, and Li}]{DBLP:journals/corr/abs-2409-14051}
Tongxuan Liu, Xingyu Wang, Weizhe Huang, Wenjiang Xu, Yuting Zeng, Lei Jiang, Hailong Yang, and Jing Li. 2024{\natexlab{a}}.
\newblock \href {https://doi.org/10.48550/ARXIV.2409.14051} {Groupdebate: Enhancing the efficiency of multi-agent debate using group discussion}.
\newblock \emph{CoRR}, abs/2409.14051.

\bibitem[{Liu et~al.(2024{\natexlab{b}})Liu, Wang, Wang, Xie, Qiu, Dang, Du, Chen, Yang, and Qian}]{DBLP:conf/nips/LiuWWXQDDCYQ24}
Wei Liu, Chenxi Wang, Yifei Wang, Zihao Xie, Rennai Qiu, Yufan Dang, Zhuoyun Du, Weize Chen, Cheng Yang, and Chen Qian. 2024{\natexlab{b}}.
\newblock \href {http://papers.nips.cc/paper\_files/paper/2024/hash/0534abc9e6db91683d82186ef0d68202-Abstract-Conference.html} {Autonomous agents for collaborative task under information asymmetry}.
\newblock In \emph{Advances in Neural Information Processing Systems 38: Annual Conference on Neural Information Processing Systems 2024, NeurIPS 2024, Vancouver, BC, Canada, December 10 - 15, 2024}.

\bibitem[{OpenAI et~al.(2024)OpenAI, :, Hurst, Lerer, Goucher, Perelman, Ramesh, Clark, Ostrow, Welihinda, Hayes, Radford, Mądry, Baker-Whitcomb, Beutel, Borzunov, Carney, Chow, Kirillov, Nichol, Paino, Renzin, Passos, Kirillov, Christakis, Conneau, Kamali, Jabri, Moyer, Tam, Crookes, Tootoochian, Tootoonchian, Kumar, Vallone, Karpathy, Braunstein, Cann, Codispoti, Galu, Kondrich, Tulloch, Mishchenko, Baek, Jiang, Pelisse, Woodford, Gosalia, Dhar, Pantuliano, Nayak, Oliver, Zoph, Ghorbani, Leimberger, Rossen, Sokolowsky, Wang, Zweig, Hoover, Samic, McGrew, Spero, Giertler, Cheng, Lightcap, Walkin, Quinn, Guarraci, Hsu, Kellogg, Eastman, Lugaresi, Wainwright, Bassin, Hudson, Chu, Nelson, Li, Shern, Conger, Barette, Voss, Ding, Lu, Zhang, Beaumont, Hallacy, Koch, Gibson, Kim, Choi, McLeavey, Hesse, Fischer, Winter, Czarnecki, Jarvis, Wei, Koumouzelis, Sherburn, Kappler, Levin, Levy, Carr, Farhi, Mely, Robinson, Sasaki, Jin, Valladares, Tsipras, Li, Nguyen, Findlay, Oiwoh, Wong, Asdar, Proehl, Yang, Antonow,
  Kramer, Peterson, Sigler, Wallace, Brevdo, Mays, Khorasani, Such, Raso, Zhang, von Lohmann, Sulit, Goh, Oden, Salmon, Starace, Brockman, Salman, Bao, Hu, Wong, Wang, Schmidt, Whitney, Jun, Kirchner, de~Oliveira~Pinto, Ren, Chang, Chung, Kivlichan, O'Connell, O'Connell, Osband, Silber, Sohl, Okuyucu, Lan, Kostrikov, Sutskever, Kanitscheider, Gulrajani, Coxon, Menick, Pachocki, Aung, Betker, Crooks, Lennon, Kiros, Leike, Park, Kwon, Phang, Teplitz, Wei, Wolfe, Chen, Harris, Varavva, Lee, Shieh, Lin, Yu, Weng, Tang, Yu, Jang, Candela, Beutler, Landers, Parish, Heidecke, Schulman, Lachman, McKay, Uesato, Ward, Kim, Huizinga, Sitkin, Kraaijeveld, Gross, Kaplan, Snyder, Achiam, Jiao, Lee, Zhuang, Harriman, Fricke, Hayashi, Singhal, Shi, Karthik, Wood, Rimbach, Hsu, Nguyen, Gu-Lemberg, Button, Liu, Howe, Muthukumar, Luther, Ahmad, Kai, Itow, Workman, Pathak, Chen, Jing, Guy, Fedus, Zhou, Mamitsuka, Weng, McCallum, Held, Ouyang, Feuvrier, Zhang, Kondraciuk, Kaiser, Hewitt, Metz, Doshi, Aflak, Simens, Boyd,
  Thompson, Dukhan, Chen, Gray, Hudnall, Zhang, Aljubeh, Litwin, Zeng, Johnson, Shetty, Gupta, Shah, Yatbaz, Yang, Zhong, Glaese, Chen, Janner, Lampe, Petrov, Wu, Wang, Fradin, Pokrass, Castro, de~Castro, Pavlov, Brundage, Wang, Khan, Murati, Bavarian, Lin, Yesildal, Soto, Gimelshein, Cone, Staudacher, Summers, LaFontaine, Chowdhury, Ryder, Stathas, Turley, Tezak, Felix, Kudige, Keskar, Deutsch, Bundick, Puckett, Nachum, Okelola, Boiko, Murk, Jaffe, Watkins, Godement, Campbell-Moore, Chao, McMillan, Belov, Su, Bak, Bakkum, Deng, Dolan, Hoeschele, Welinder, Tillet, Pronin, Tillet, Dhariwal, Yuan, Dias, Lim, Arora, Troll, Lin, Lopes, Puri, Miyara, Leike, Gaubert, Zamani, Wang, Donnelly, Honsby, Smith, Sahai, Ramchandani, Huet, Carmichael, Zellers, Chen, Chen, Nigmatullin, Cheu, Jain, Altman, Schoenholz, Toizer, Miserendino, Agarwal, Culver, Ethersmith, Gray, Grove, Metzger, Hermani, Jain, Zhao, Wu, Jomoto, Wu, Shuaiqi, Xia, Phene, Papay, Narayanan, Coffey, Lee, Hall, Balaji, Broda, Stramer, Xu, Gogineni,
  Christianson, Sanders, Patwardhan, Cunninghman, Degry, Dimson, Raoux, Shadwell, Zheng, Underwood, Markov, Sherbakov, Rubin, Stasi, Kaftan, Heywood, Peterson, Walters, Eloundou, Qi, Moeller, Monaco, Kuo, Fomenko, Chang, Zheng, Zhou, Manassra, Sheu, Zaremba, Patil, Qian, Kim, Cheng, Zhang, He, Zhang, Jin, Dai, and Malkov}]{openai2024gpt4ocard}
OpenAI, :, Aaron Hurst, Adam Lerer, Adam~P. Goucher, Adam Perelman, Aditya Ramesh, Aidan Clark, AJ~Ostrow, Akila Welihinda, Alan Hayes, Alec Radford, Aleksander Mądry, Alex Baker-Whitcomb, Alex Beutel, Alex Borzunov, Alex Carney, Alex Chow, Alex Kirillov, Alex Nichol, Alex Paino, Alex Renzin, Alex~Tachard Passos, Alexander Kirillov, Alexi Christakis, Alexis Conneau, Ali Kamali, Allan Jabri, Allison Moyer, Allison Tam, Amadou Crookes, Amin Tootoochian, Amin Tootoonchian, Ananya Kumar, Andrea Vallone, Andrej Karpathy, Andrew Braunstein, Andrew Cann, Andrew Codispoti, Andrew Galu, Andrew Kondrich, Andrew Tulloch, Andrey Mishchenko, Angela Baek, Angela Jiang, Antoine Pelisse, Antonia Woodford, Anuj Gosalia, Arka Dhar, Ashley Pantuliano, Avi Nayak, Avital Oliver, Barret Zoph, Behrooz Ghorbani, Ben Leimberger, Ben Rossen, Ben Sokolowsky, Ben Wang, Benjamin Zweig, Beth Hoover, Blake Samic, Bob McGrew, Bobby Spero, Bogo Giertler, Bowen Cheng, Brad Lightcap, Brandon Walkin, Brendan Quinn, Brian Guarraci, Brian Hsu,
  Bright Kellogg, Brydon Eastman, Camillo Lugaresi, Carroll Wainwright, Cary Bassin, Cary Hudson, Casey Chu, Chad Nelson, Chak Li, Chan~Jun Shern, Channing Conger, Charlotte Barette, Chelsea Voss, Chen Ding, Cheng Lu, Chong Zhang, Chris Beaumont, Chris Hallacy, Chris Koch, Christian Gibson, Christina Kim, Christine Choi, Christine McLeavey, Christopher Hesse, Claudia Fischer, Clemens Winter, Coley Czarnecki, Colin Jarvis, Colin Wei, Constantin Koumouzelis, Dane Sherburn, Daniel Kappler, Daniel Levin, Daniel Levy, David Carr, David Farhi, David Mely, David Robinson, David Sasaki, Denny Jin, Dev Valladares, Dimitris Tsipras, Doug Li, Duc~Phong Nguyen, Duncan Findlay, Edede Oiwoh, Edmund Wong, Ehsan Asdar, Elizabeth Proehl, Elizabeth Yang, Eric Antonow, Eric Kramer, Eric Peterson, Eric Sigler, Eric Wallace, Eugene Brevdo, Evan Mays, Farzad Khorasani, Felipe~Petroski Such, Filippo Raso, Francis Zhang, Fred von Lohmann, Freddie Sulit, Gabriel Goh, Gene Oden, Geoff Salmon, Giulio Starace, Greg Brockman, Hadi
  Salman, Haiming Bao, Haitang Hu, Hannah Wong, Haoyu Wang, Heather Schmidt, Heather Whitney, Heewoo Jun, Hendrik Kirchner, Henrique~Ponde de~Oliveira~Pinto, Hongyu Ren, Huiwen Chang, Hyung~Won Chung, Ian Kivlichan, Ian O'Connell, Ian O'Connell, Ian Osband, Ian Silber, Ian Sohl, Ibrahim Okuyucu, Ikai Lan, Ilya Kostrikov, Ilya Sutskever, Ingmar Kanitscheider, Ishaan Gulrajani, Jacob Coxon, Jacob Menick, Jakub Pachocki, James Aung, James Betker, James Crooks, James Lennon, Jamie Kiros, Jan Leike, Jane Park, Jason Kwon, Jason Phang, Jason Teplitz, Jason Wei, Jason Wolfe, Jay Chen, Jeff Harris, Jenia Varavva, Jessica~Gan Lee, Jessica Shieh, Ji~Lin, Jiahui Yu, Jiayi Weng, Jie Tang, Jieqi Yu, Joanne Jang, Joaquin~Quinonero Candela, Joe Beutler, Joe Landers, Joel Parish, Johannes Heidecke, John Schulman, Jonathan Lachman, Jonathan McKay, Jonathan Uesato, Jonathan Ward, Jong~Wook Kim, Joost Huizinga, Jordan Sitkin, Jos Kraaijeveld, Josh Gross, Josh Kaplan, Josh Snyder, Joshua Achiam, Joy Jiao, Joyce Lee, Juntang
  Zhuang, Justyn Harriman, Kai Fricke, Kai Hayashi, Karan Singhal, Katy Shi, Kavin Karthik, Kayla Wood, Kendra Rimbach, Kenny Hsu, Kenny Nguyen, Keren Gu-Lemberg, Kevin Button, Kevin Liu, Kiel Howe, Krithika Muthukumar, Kyle Luther, Lama Ahmad, Larry Kai, Lauren Itow, Lauren Workman, Leher Pathak, Leo Chen, Li~Jing, Lia Guy, Liam Fedus, Liang Zhou, Lien Mamitsuka, Lilian Weng, Lindsay McCallum, Lindsey Held, Long Ouyang, Louis Feuvrier, Lu~Zhang, Lukas Kondraciuk, Lukasz Kaiser, Luke Hewitt, Luke Metz, Lyric Doshi, Mada Aflak, Maddie Simens, Madelaine Boyd, Madeleine Thompson, Marat Dukhan, Mark Chen, Mark Gray, Mark Hudnall, Marvin Zhang, Marwan Aljubeh, Mateusz Litwin, Matthew Zeng, Max Johnson, Maya Shetty, Mayank Gupta, Meghan Shah, Mehmet Yatbaz, Meng~Jia Yang, Mengchao Zhong, Mia Glaese, Mianna Chen, Michael Janner, Michael Lampe, Michael Petrov, Michael Wu, Michele Wang, Michelle Fradin, Michelle Pokrass, Miguel Castro, Miguel Oom~Temudo de~Castro, Mikhail Pavlov, Miles Brundage, Miles Wang, Minal
  Khan, Mira Murati, Mo~Bavarian, Molly Lin, Murat Yesildal, Nacho Soto, Natalia Gimelshein, Natalie Cone, Natalie Staudacher, Natalie Summers, Natan LaFontaine, Neil Chowdhury, Nick Ryder, Nick Stathas, Nick Turley, Nik Tezak, Niko Felix, Nithanth Kudige, Nitish Keskar, Noah Deutsch, Noel Bundick, Nora Puckett, Ofir Nachum, Ola Okelola, Oleg Boiko, Oleg Murk, Oliver Jaffe, Olivia Watkins, Olivier Godement, Owen Campbell-Moore, Patrick Chao, Paul McMillan, Pavel Belov, Peng Su, Peter Bak, Peter Bakkum, Peter Deng, Peter Dolan, Peter Hoeschele, Peter Welinder, Phil Tillet, Philip Pronin, Philippe Tillet, Prafulla Dhariwal, Qiming Yuan, Rachel Dias, Rachel Lim, Rahul Arora, Rajan Troll, Randall Lin, Rapha~Gontijo Lopes, Raul Puri, Reah Miyara, Reimar Leike, Renaud Gaubert, Reza Zamani, Ricky Wang, Rob Donnelly, Rob Honsby, Rocky Smith, Rohan Sahai, Rohit Ramchandani, Romain Huet, Rory Carmichael, Rowan Zellers, Roy Chen, Ruby Chen, Ruslan Nigmatullin, Ryan Cheu, Saachi Jain, Sam Altman, Sam Schoenholz, Sam
  Toizer, Samuel Miserendino, Sandhini Agarwal, Sara Culver, Scott Ethersmith, Scott Gray, Sean Grove, Sean Metzger, Shamez Hermani, Shantanu Jain, Shengjia Zhao, Sherwin Wu, Shino Jomoto, Shirong Wu, Shuaiqi, Xia, Sonia Phene, Spencer Papay, Srinivas Narayanan, Steve Coffey, Steve Lee, Stewart Hall, Suchir Balaji, Tal Broda, Tal Stramer, Tao Xu, Tarun Gogineni, Taya Christianson, Ted Sanders, Tejal Patwardhan, Thomas Cunninghman, Thomas Degry, Thomas Dimson, Thomas Raoux, Thomas Shadwell, Tianhao Zheng, Todd Underwood, Todor Markov, Toki Sherbakov, Tom Rubin, Tom Stasi, Tomer Kaftan, Tristan Heywood, Troy Peterson, Tyce Walters, Tyna Eloundou, Valerie Qi, Veit Moeller, Vinnie Monaco, Vishal Kuo, Vlad Fomenko, Wayne Chang, Weiyi Zheng, Wenda Zhou, Wesam Manassra, Will Sheu, Wojciech Zaremba, Yash Patil, Yilei Qian, Yongjik Kim, Youlong Cheng, Yu~Zhang, Yuchen He, Yuchen Zhang, Yujia Jin, Yunxing Dai, and Yury Malkov. 2024.
\newblock \href {https://arxiv.org/abs/2410.21276} {Gpt-4o system card}.
\newblock \emph{Preprint}, arXiv:2410.21276.

\bibitem[{Piao et~al.(2025)Piao, Yan, Zhang, Li, Yan, Lan, Lu, Zheng, Wang, Zhou, Gao, Xu, Zhang, Rong, Su, and Li}]{DBLP:journals/corr/abs-2502-08691}
Jinghua Piao, Yuwei Yan, Jun Zhang, Nian Li, Junbo Yan, Xiaochong Lan, Zhihong Lu, Zhiheng Zheng, Jing~Yi Wang, Di~Zhou, Chen Gao, Fengli Xu, Fang Zhang, Ke~Rong, Jun Su, and Yong Li. 2025.
\newblock \href {https://doi.org/10.48550/ARXIV.2502.08691} {Agentsociety: Large-scale simulation of llm-driven generative agents advances understanding of human behaviors and society}.
\newblock \emph{CoRR}, abs/2502.08691.

\bibitem[{Qian et~al.(2024)Qian, Xie, Wang, Liu, Dang, Du, Chen, Yang, Liu, and Sun}]{DBLP:journals/corr/abs-2406-07155}
Chen Qian, Zihao Xie, Yifei Wang, Wei Liu, Yufan Dang, Zhuoyun Du, Weize Chen, Cheng Yang, Zhiyuan Liu, and Maosong Sun. 2024.
\newblock \href {https://doi.org/10.48550/ARXIV.2406.07155} {Scaling large-language-model-based multi-agent collaboration}.
\newblock \emph{CoRR}, abs/2406.07155.

\bibitem[{Reimers and Gurevych(2019)}]{DBLP:conf/emnlp/ReimersG19}
Nils Reimers and Iryna Gurevych. 2019.
\newblock \href {https://doi.org/10.18653/V1/D19-1410} {Sentence-bert: Sentence embeddings using siamese bert-networks}.
\newblock In \emph{Proceedings of the 2019 Conference on Empirical Methods in Natural Language Processing and the 9th International Joint Conference on Natural Language Processing, {EMNLP-IJCNLP} 2019, Hong Kong, China, November 3-7, 2019}, pages 3980--3990. Association for Computational Linguistics.

\bibitem[{Samuel et~al.(2024)Samuel, Zou, Zhou, Chaudhari, Kalyan, Rajpurohit, Deshpande, Narasimhan, and Murahari}]{DBLP:journals/corr/abs-2407-18416}
Vinay Samuel, Henry~Peng Zou, Yue Zhou, Shreyas Chaudhari, Ashwin Kalyan, Tanmay Rajpurohit, Ameet Deshpande, Karthik Narasimhan, and Vishvak Murahari. 2024.
\newblock \href {https://doi.org/10.48550/ARXIV.2407.18416} {Personagym: Evaluating persona agents and llms}.
\newblock \emph{CoRR}, abs/2407.18416.

\bibitem[{Surowiecki(2004)}]{Surowiecki2004Wisdom}
James Surowiecki. 2004.
\newblock \emph{{The Wisdom of Crowds}}.
\newblock Doubleday, New York.
\newblock Subtitle: Why the Many Are Smarter Than the Few and How Collective Wisdom Shapes Business, Economies, Societies and Nations -- included subtitle in note as it's long, adjust if needed.

\bibitem[{Team et~al.(2025)Team, Du, Gao, Xing, Jiang, Chen, Li, Xiao, Du, Liao, Tang, Wang, Zhang, Yuan, Lu, Tang, Sung, Wei, Lai, Guo, Zhu, Ding, Hu, Yang, Zhang, Yao, Zhao, Lu, Li, Yu, Gao, Zheng, Yuan, Chen, Guo, Su, Wang, Zhao, Zhang, Liu, Yan, Wu, Shi, Ye, Yu, Dong, Zhang, Ma, Pan, Gong, Liu, Ma, Wei, Cao, Huang, Jiang, Gao, Xiong, He, Huang, Wu, He, Wei, Jia, Wu, Xu, Zu, Zhou, Pan, Charles, Li, Hu, Liu, Chen, Wang, Liu, Qin, Liu, Yang, Bao, Du, Wu, Wang, Zhou, Wang, Li, Zhu, Zhang, Wang, Yang, Huang, Huang, Xu, and Yang}]{DBLP:journals/corr/abs-2501-12599}
Kimi Team, Angang Du, Bofei Gao, Bowei Xing, Changjiu Jiang, Cheng Chen, Cheng Li, Chenjun Xiao, Chenzhuang Du, Chonghua Liao, Chuning Tang, Congcong Wang, Dehao Zhang, Enming Yuan, Enzhe Lu, Fengxiang Tang, Flood Sung, Guangda Wei, Guokun Lai, Haiqing Guo, Han Zhu, Hao Ding, Hao Hu, Hao Yang, Hao Zhang, Haotian Yao, Haotian Zhao, Haoyu Lu, Haoze Li, Haozhen Yu, Hongcheng Gao, Huabin Zheng, Huan Yuan, Jia Chen, Jianhang Guo, Jianlin Su, Jianzhou Wang, Jie Zhao, Jin Zhang, Jingyuan Liu, Junjie Yan, Junyan Wu, Lidong Shi, Ling Ye, Longhui Yu, Mengnan Dong, Neo Zhang, Ningchen Ma, Qiwei Pan, Qucheng Gong, Shaowei Liu, Shengling Ma, Shupeng Wei, Sihan Cao, Siying Huang, Tao Jiang, Weihao Gao, Weimin Xiong, Weiran He, Weixiao Huang, Wenhao Wu, Wenyang He, Xianghui Wei, Xianqing Jia, Xingzhe Wu, Xinran Xu, Xinxing Zu, Xinyu Zhou, Xuehai Pan, Y.~Charles, Yang Li, Yangyang Hu, Yangyang Liu, Yanru Chen, Yejie Wang, Yibo Liu, Yidao Qin, Yifeng Liu, Ying Yang, Yiping Bao, Yulun Du, Yuxin Wu, Yuzhi Wang, Zaida Zhou,
  Zhaoji Wang, Zhaowei Li, Zhen Zhu, Zheng Zhang, Zhexu Wang, Zhilin Yang, Zhiqi Huang, Zihao Huang, Ziyao Xu, and Zonghan Yang. 2025.
\newblock \href {https://doi.org/10.48550/ARXIV.2501.12599} {Kimi k1.5: Scaling reinforcement learning with llms}.
\newblock \emph{CoRR}, abs/2501.12599.

\bibitem[{Wang et~al.(2024{\natexlab{a}})Wang, Wang, Athiwaratkun, Zhang, and Zou}]{DBLP:journals/corr/abs-2406-04692}
Junlin Wang, Jue Wang, Ben Athiwaratkun, Ce~Zhang, and James Zou. 2024{\natexlab{a}}.
\newblock \href {https://doi.org/10.48550/ARXIV.2406.04692} {Mixture-of-agents enhances large language model capabilities}.
\newblock \emph{CoRR}, abs/2406.04692.

\bibitem[{Wang et~al.(2024{\natexlab{b}})Wang, Wang, Su, Tong, and Song}]{DBLP:conf/acl/WangWSTS24}
Qineng Wang, Zihao Wang, Ying Su, Hanghang Tong, and Yangqiu Song. 2024{\natexlab{b}}.
\newblock \href {https://doi.org/10.18653/V1/2024.ACL-LONG.331} {Rethinking the bounds of {LLM} reasoning: Are multi-agent discussions the key?}
\newblock In \emph{Proceedings of the 62nd Annual Meeting of the Association for Computational Linguistics (Volume 1: Long Papers), {ACL} 2024, Bangkok, Thailand, August 11-16, 2024}, pages 6106--6131. Association for Computational Linguistics.

\bibitem[{Wang et~al.(2024{\natexlab{c}})Wang, Yu, Zhang, Qi, Sap, Bisk, Neubig, and Zhu}]{DBLP:conf/acl/WangYZQSBN024}
Ruiyi Wang, Haofei Yu, Wenxin~Sharon Zhang, Zhengyang Qi, Maarten Sap, Yonatan Bisk, Graham Neubig, and Hao Zhu. 2024{\natexlab{c}}.
\newblock \href {https://doi.org/10.18653/V1/2024.ACL-LONG.698} {Sotopia-{\(\pi\)}: Interactive learning of socially intelligent language agents}.
\newblock In \emph{Proceedings of the 62nd Annual Meeting of the Association for Computational Linguistics (Volume 1: Long Papers), {ACL} 2024, Bangkok, Thailand, August 11-16, 2024}, pages 12912--12940. Association for Computational Linguistics.

\bibitem[{Wang et~al.(2024{\natexlab{d}})Wang, Ma, Zhang, Ni, Chandra, Guo, Ren, Arulraj, He, Jiang, Li, Ku, Wang, Zhuang, Fan, Yue, and Chen}]{DBLP:conf/nips/WangMZNCGRAHJLK24}
Yubo Wang, Xueguang Ma, Ge~Zhang, Yuansheng Ni, Abhranil Chandra, Shiguang Guo, Weiming Ren, Aaran Arulraj, Xuan He, Ziyan Jiang, Tianle Li, Max Ku, Kai Wang, Alex Zhuang, Rongqi Fan, Xiang Yue, and Wenhu Chen. 2024{\natexlab{d}}.
\newblock \href {http://papers.nips.cc/paper\_files/paper/2024/hash/ad236edc564f3e3156e1b2feafb99a24-Abstract-Datasets\_and\_Benchmarks\_Track.html} {Mmlu-pro: {A} more robust and challenging multi-task language understanding benchmark}.
\newblock In \emph{Advances in Neural Information Processing Systems 38: Annual Conference on Neural Information Processing Systems 2024, NeurIPS 2024, Vancouver, BC, Canada, December 10 - 15, 2024}.

\bibitem[{Xu et~al.(2024{\natexlab{a}})Xu, Wang, Shi, Ding, Jing, Fang, Bai, Liu, Yu, Li, Luo, Yin, Yin, Chen, and Song}]{DBLP:conf/emnlp/Xu0S0JFBLYLLYYC24}
Baixuan Xu, Weiqi Wang, Haochen Shi, Wenxuan Ding, Huihao Jing, Tianqing Fang, Jiaxin Bai, Xin Liu, Changlong Yu, Zheng Li, Chen Luo, Qingyu Yin, Bing Yin, Long Chen, and Yangqiu Song. 2024{\natexlab{a}}.
\newblock \href {https://aclanthology.org/2024.emnlp-main.446} {{MIND:} multimodal shopping intention distillation from large vision-language models for e-commerce purchase understanding}.
\newblock In \emph{Proceedings of the 2024 Conference on Empirical Methods in Natural Language Processing, {EMNLP} 2024, Miami, FL, USA, November 12-16, 2024}, pages 7800--7815. Association for Computational Linguistics.

\bibitem[{Xu et~al.(2023)Xu, Yang, Lin, Wang, Zhou, Zhang, and Mao}]{DBLP:journals/corr/abs-2305-14688}
Benfeng Xu, An~Yang, Junyang Lin, Quan Wang, Chang Zhou, Yongdong Zhang, and Zhendong Mao. 2023.
\newblock \href {https://doi.org/10.48550/ARXIV.2305.14688} {Expertprompting: Instructing large language models to be distinguished experts}.
\newblock \emph{CoRR}, abs/2305.14688.

\bibitem[{Xu et~al.(2024{\natexlab{b}})Xu, Song, Li, Tang, Jain, Bao, Wang, Zhou, Guo, Cao, Yang, Lu, Martin, Su, Maben, Mehta, Chi, Jang, Xie, Zhou, and Neubig}]{DBLP:journals/corr/abs-2412-14161}
Frank~F. Xu, Yufan Song, Boxuan Li, Yuxuan Tang, Kritanjali Jain, Mengxue Bao, Zora~Z. Wang, Xuhui Zhou, Zhitong Guo, Murong Cao, Mingyang Yang, Hao~Yang Lu, Amaad Martin, Zhe Su, Leander Maben, Raj Mehta, Wayne Chi, Lawrence~Keunho Jang, Yiqing Xie, Shuyan Zhou, and Graham Neubig. 2024{\natexlab{b}}.
\newblock \href {https://doi.org/10.48550/ARXIV.2412.14161} {Theagentcompany: Benchmarking {LLM} agents on consequential real world tasks}.
\newblock \emph{CoRR}, abs/2412.14161.

\bibitem[{Xu et~al.(2024{\natexlab{c}})Xu, Ping, Wu, McAfee, Zhu, Liu, Subramanian, Bakhturina, Shoeybi, and Catanzaro}]{DBLP:conf/iclr/0008PWM0LSBSC24}
Peng Xu, Wei Ping, Xianchao Wu, Lawrence McAfee, Chen Zhu, Zihan Liu, Sandeep Subramanian, Evelina Bakhturina, Mohammad Shoeybi, and Bryan Catanzaro. 2024{\natexlab{c}}.
\newblock \href {https://openreview.net/forum?id=xw5nxFWMlo} {Retrieval meets long context large language models}.
\newblock In \emph{The Twelfth International Conference on Learning Representations, {ICLR} 2024, Vienna, Austria, May 7-11, 2024}. OpenReview.net.

\bibitem[{Yang et~al.(2025)Yang, Yu, Li, Liu, Huang, Huang, Jiang, Tu, Zhang, Zhou, Lin, Dang, Yang, Yu, Li, Sun, Zhu, Men, He, Xu, Yin, Yu, Qiu, Ren, Yang, Li, Xu, and Zhang}]{DBLP:journals/corr/abs-2501-15383}
An~Yang, Bowen Yu, Chengyuan Li, Dayiheng Liu, Fei Huang, Haoyan Huang, Jiandong Jiang, Jianhong Tu, Jianwei Zhang, Jingren Zhou, Junyang Lin, Kai Dang, Kexin Yang, Le~Yu, Mei Li, Minmin Sun, Qin Zhu, Rui Men, Tao He, Weijia Xu, Wenbiao Yin, Wenyuan Yu, Xiafei Qiu, Xingzhang Ren, Xinlong Yang, Yong Li, Zhiying Xu, and Zipeng Zhang. 2025.
\newblock \href {https://doi.org/10.48550/ARXIV.2501.15383} {Qwen2.5-1m technical report}.
\newblock \emph{CoRR}, abs/2501.15383.

\bibitem[{Yang et~al.(2024)Yang, Zhang, Zheng, Jiang, Gan, Wang, Ling, Chen, Ma, Dong, Gupta, Hu, Yin, Li, Jia, Wang, Ghanem, Lu, Lu, Ouyang, Qiao, Torr, and Shao}]{DBLP:journals/corr/abs-2411-11581}
Ziyi Yang, Zaibin Zhang, Zirui Zheng, Yuxian Jiang, Ziyue Gan, Zhiyu Wang, Zijian Ling, Jinsong Chen, Martz Ma, Bowen Dong, Prateek Gupta, Shuyue Hu, Zhenfei Yin, Guohao Li, Xu~Jia, Lijun Wang, Bernard Ghanem, Huchuan Lu, Chaochao Lu, Wanli Ouyang, Yu~Qiao, Philip Torr, and Jing Shao. 2024.
\newblock \href {https://doi.org/10.48550/ARXIV.2411.11581} {{OASIS:} open agent social interaction simulations with one million agents}.
\newblock \emph{CoRR}, abs/2411.11581.

\end{thebibliography}

\appendix
\newpage
\section{Agent Communication Algorithm}
\label{app:algo}
In this section, we provide our detailed algorithm for inter-agent communication protocol and its corresponding notation table in below.

\begin{algorithm}
\caption{Communication Mechanism}\label{appendix:Communication Algorithm}
\begin{algorithmic}
\Procedure{Collaboration}{$\mathcal{Q}, \mathcal{S},\mathcal{M}_n$}
    \For{$\mathcal{A}_i$ in $\mathcal{M}_n$}
        \If{i = n}
            \State $\mathcal{Y} \leftarrow \mathcal{A}_{n}(\mathcal{Q},\mathcal{S},\mathcal{A}_{1},...,\mathcal{A}^{f}_{n-1})$
            \State \Return $\mathcal{Y}$
        \ElsIf{i = 1}
            \State $\mathcal{Y}\leftarrow \mathcal{A}_{1}(\mathcal{Q},\mathcal{S})$
        \Else
            \State $\mathcal{Y} \leftarrow \mathcal{A}_{i}(\mathcal{Q},\mathcal{S},\mathcal{A}_{1},...,\mathcal{A}^{f}_{i-1})$  
        \EndIf
    \EndFor
\EndProcedure
\end{algorithmic}
\end{algorithm}

\begin{table}[ht]
\centering
\small
\resizebox{\linewidth}{!}{%
\begin{tabular}{cc}
\hline
Symbol & Meaning \\
\hline
$\mathcal{A}_i$ & The output without rationale of agent $\mathcal{A}_i$ \\
$\mathcal{A}^f_i$ & Full output with rationale of agent $\mathcal{A}_i$ \\
$\mathcal{Q}$ & Input question \\
$\mathcal{S}$ & The candidate answers of the question \\
$\mathcal{Y}$ & The final answer of the system
\\
\hline
\end{tabular}
}
\caption{Notation used in Algorithm \ref{appendix:Communication Algorithm}}
\end{table}

\begin{figure*}[h]
\section{Role System Prompt}
\label{app:system_prompt}
\begin{tcolorbox}[title={System Prompt}, colback = cGrey, colframe = black,  coltitle=white,fonttitle=\bfseries\small, center title,fontupper=\small,fontlower=\small]
\begin{center}
[ROLE ASSIGNMENT]    
\end{center}
You are a \{title\} specializing in \{domain\}.\\
Your professional responsibility is to \{duty\}.\\
IMPORTANT: Think and respond EXACTLY as a real \{title\} in \{domain\} would.\\
Use terminology, methods, and perspectives specific to your professional field.
\end{tcolorbox}
\end{figure*}

\begin{figure*}[h]
\begin{tcolorbox}[title={User Prompt}, colback = cGrey, colframe = black,  coltitle=white,fonttitle=\bfseries\small, center title,fontupper=\small,fontlower=\small]
Previous discussion:    \{message\_hist\}
PROBLEM TO SOLVE: {problem}
RESPONSE INSTRUCTIONS:
1. Begin with: "As a \{title\} in \{domain\}, I..."
2. Analyze the problem using your professional expertise
3. Provide your expert recommendation
4. End with: "My answer is \\boxed\{\{X\}\}" where X is the answer index

REQUIREMENTS:
- Maintain your \{title\} perspective throughout
- Use terminology from \{domain\}
- Keep response under 150 words
- Your answer MUST be in \\boxed\{\{\}\} format

Remember: You are a \{title\}, not an AI assistant. Think and respond accordingly.
\end{tcolorbox}
\end{figure*}

\begin{figure*}[b]
\section{Expert Generation Prompts}
\label{app:role_prompts_gen}
\label{app:Expert Generation Prompts}
\subsection{Primary Expert Generation Prompts}
\centering
\begin{tcolorbox}[title={Prompt for Structured Workflow Expert Generation}, colback = cBlue_1!10, colframe = cBlue_7,  coltitle=white,fonttitle=\bfseries\small, center title,fontupper=\small,fontlower=\small]
\textbf{Variables:} \{Domain\}
\tcblower
\textbf{Prompt:}
Generate me an expert group in {Domain} domain of size three, assigning them roles of solver, critic and coordinator together with their detailed responsibilities.
\end{tcolorbox}
\end{figure*}

\begin{figure*}[h]
\begin{tcolorbox}[title={Prompt for Diversity-Driven Expert Generation}, colback = cBlue_1!10, colframe = cBlue_7,  coltitle=white,fonttitle=\bfseries\small, center title,fontupper=\small,fontlower=\small]
\textbf{Variables:} \{Domain\}
\tcblower
\textbf{Prompt:}
Generate an expert group of size 3 in the {Domain} domain, each specializing in a distinct sub-domain of {Domain}. Provide a detailed configuration for each expert, including their role and responsibility, ensuring that their roles are complementary and collectively form a balanced, high-functioning team capable of addressing complex challenges in the domain.
For example, an expert in a sub-domain of business could be ``Global Compliance Architect''.
\end{tcolorbox}
\end{figure*}

\begin{figure*}
\subsection{Expert Augmentation Process}
\begin{tcolorbox}[title={Prompt for Structured Workflow Expert Augmentation}, colback = cBlue_1!10, colframe = cBlue_7,  coltitle=white,fonttitle=\bfseries\small, center title,fontupper=\small,fontlower=\small]
\textbf{Variables:} \{Domain\},\{System Size\},\{Group Description of Size 3\}
\tcblower
\textbf{Prompt:}
Here is a expert group configuration in {Domain} domain of size 3: {Group Description of Size 3}.\\
Please augment the group size to {System Size} by assigning new experts with roles of solver, critic, strategist and coordinator.
Output your configuration following the format of the given group configuration.
\end{tcolorbox}
\end{figure*}

\begin{figure*}
\begin{tcolorbox}[title={Prompt for Diversity-Driven Expert Augmentation}, colback = cBlue_1!10, colframe = cBlue_7,  coltitle=white,fonttitle=\bfseries\small, center title,fontupper=\small,fontlower=\small]
\textbf{Variables:} \{Domain\},\{System Size\},\{Group Description of Size 3\}
\tcblower
\textbf{Prompt:}
Here is a expert group configuration in {Domain} domain of size 3: {Group Description of Size 3}.\\
Please augment the group size to {System Size} by assigning new experts with roles of expert in other sub-domains in {Domain} together with their responsibilities.
Output your configuration following the format of the given group configuration.
\end{tcolorbox}
\end{figure*}
\label{app:scaling_role_prompts}

\begin{figure*}[h]
\section{Social Group Role Examples}
\label{Appendix:Social Group Role Examples}
\begin{tcolorbox}[title={Math Group of 3}, colback = cGrey, colframe = black,  coltitle=white,fonttitle=\bfseries\small, center title,fontupper=\small,fontlower=\small]
\begin{center}
\MakeUppercase{\romannum{1}}.  Differential Topologist
\end{center}
Responsibilities:\\
1. Analyze manifold embeddings using Whitney's conditions\\
2. Verify cobordism relations through Morse homology\\
3. Calculate characteristic classes via Čech-de Rham complexes\\

\begin{center}
\MakeUppercase{\romannum{2}}. Proof Metrologist
\end{center}
Responsibilities:\\
1. Audit natural deduction derivations for intuitionistic consistency\\
2. Identify unstated ZFC dependencies\\
3. Verify category-theoretic diagram commutativity\\

\begin{center}
\MakeUppercase{\romannum{3}}. 
Spectral Synthesizer
\end{center}
Responsibilities:\\
1. Decompose operator algebras using K-theory invariants\\
2. Construct Gelfand-Naimark-Segal representations\\
3. Analyze C*-algebra extension groups\\
\end{tcolorbox}
\end{figure*}

\begin{figure*}[h]
\begin{tcolorbox}[title={Math Group of 3}, colback = cGrey, colframe = black,  coltitle=white,fonttitle=\bfseries\small, center title,fontupper=\small,fontlower=\small]
\begin{center}
\MakeUppercase{\romannum{1}}.  Solver
\end{center}
Responsibilities:\\
execute core problem analysis using mathematical principles, formulate key equations, and establish foundational solution components with logical progression.\\

\begin{center}
\MakeUppercase{\romannum{2}}. Critic
\end{center}
Responsibilities:\\
Analyze solution structure for conceptual consistency, identify invalid logical leaps, and verify fundamental mathematical truth of initial assumptions.\\

\begin{center}
\MakeUppercase{\romannum{3}}. 
Coordinator
\end{center}
Responsibilities:\\
Integrate analytical components into unified framework, maintain mathematical coherence between steps, and prepare final solution presentation.\\
\end{tcolorbox}
\end{figure*}

\begin{figure*}[h]
\begin{tcolorbox}[title={Finance Group of 3}, colback = cGrey, colframe = black,  coltitle=white,fonttitle=\bfseries\small, center title,fontupper=\small,fontlower=\small]
\begin{center}
\MakeUppercase{\romannum{1}}.  Ethics \& Compliance Officer
\end{center}
Responsibilities:\\
1. Merge UNGC/SBE mapping with FTC/ASA/CAP compliance\\
2. Conduct combined PESTEL/SWOT analyses\\
3. Integrate CSR violation detection with greenwashing audits\\
4. Handle stakeholder prioritization with power-interest matrices\\
5. Develop unified compliance solutions using BIA/GVV frameworks\\

\begin{center}
\MakeUppercase{\romannum{2}}. Stakeholder Impact Strategist
\end{center}
Responsibilities:\\
1.Combine emotional valence analysis with reputational scoring\\
2.Merge Maslow's hierarchy applications with PROTECT framework\\
3.Manage supply chain/social impact predictions\\
4.Balance shareholder-stakeholder priorities\\
5.Coordinate multi-channel communication plans\\

\begin{center}
\MakeUppercase{\romannum{3}}. 
Strategic Decision Leader
\end{center}
Responsibilities:\\
1.Integrate Monte Carlo simulations with game theory models\\
2.Oversee crisis protocol development/implementation\\
3.Manage alternative scenario planning\\
4.Conduct comprehensive risk-reward analysis\\
5.Finalize violation classifications/severity gradations
\end{tcolorbox}
\end{figure*}

\begin{figure*}[h]
\begin{tcolorbox}[title={Finance Group of 3}, colback = cGrey, colframe = black,  coltitle=white,fonttitle=\bfseries\small, center title,fontupper=\small,fontlower=\small]
\begin{center}
\MakeUppercase{\romannum{1}}.  Solver
\end{center}
Responsibilities:\\
Analyze regulatory compliance requirements, develop ethical frameworks, and optimize corporate governance strategies.\\

\begin{center}
\MakeUppercase{\romannum{2}}. 
Critic
\end{center}
Responsibilities:\\
Evaluate stakeholder impact scenarios, identify compliance gaps, and verify ethical decision-making processes.\\

\begin{center}
\MakeUppercase{\romannum{3}}. 
Coordinator
\end{center}
Responsibilities:\\
Integrate global compliance standards with local operations, balance stakeholder priorities, and ensure ethical crisis management.
\end{tcolorbox}
\end{figure*}

\begin{figure*}[h]
\begin{tcolorbox}[title={Medical Group of 3}, colback = cGrey, colframe = black,  coltitle=white,fonttitle=\bfseries\small, center title,fontupper=\small,fontlower=\small]
\begin{center}
\MakeUppercase{\romannum{1}}.  Disease Control Integrator
\end{center}
Responsibilities:\\
1.Combine SEIR modeling with transmission vector mapping\\
2.Merge clinical/public health intervention analysis\\
3.Integrate prevention frameworks with treatment protocols\\
4.Conduct combined cost-effectiveness/equity assessments\\
5.Develop unified outbreak response plans\\

\begin{center}
\MakeUppercase{\romannum{2}}. Health Systems Engineer
\end{center}
Responsibilities:\\
1.Synthesize care delivery models with infrastructure analysis\\
2.Optimize vaccine protocols with screening algorithms\\
3.Manage digital health/supply chain integration\\
4.Balance individual/population health needs\\
5.Conduct pandemic preparedness simulations\\

\begin{center}
\MakeUppercase{\romannum{3}}. 
Medical Priority Strategist
\end{center}
Responsibilities:\\
1.Reconcile SDG targets with local health realities\\
2.Apply GRADE criteria to population health approaches\\
3.Design risk-stratified intervention cascades\\
4.Finalize biological plausibility/scalability assessments\\
5.Produce multi-level prevention-treatment packages
\end{tcolorbox}
\end{figure*}

\begin{figure*}[h]
\begin{tcolorbox}[title={Medical Group of 3}, colback = cGrey, colframe = black,  coltitle=white,fonttitle=\bfseries\small, center title,fontupper=\small,fontlower=\small]
\begin{center}
\MakeUppercase{\romannum{1}}.  Solver
\end{center}
Responsibilities:\\
Analyze disease patterns and treatment effectiveness, develop care protocols, and optimize clinical workflows for patient outcomes.\\

\begin{center}
\MakeUppercase{\romannum{2}}. Critic
\end{center}
Responsibilities:\\
Evaluate treatment safety and efficacy, identify gaps in care standards, and verify compliance with medical guidelines.\\

\begin{center}
\MakeUppercase{\romannum{3}}. 
Coordinator
\end{center}
Responsibilities:\\
Integrate preventive care with treatment services, manage resource allocation, and ensure continuity of care across providers.
\end{tcolorbox}
\end{figure*}

\begin{figure*}[h]
\begin{tcolorbox}[title={Law Group of 3}, colback = cGrey, colframe = black,  coltitle=white,fonttitle=\bfseries\small, center title,fontupper=\small,fontlower=\small]
\begin{center}
\MakeUppercase{\romannum{1}}.  Contract Architect
\end{center}
Responsibilities:\\
1.Analyze UCC provisions vs common law principles\\
2.Identify material breach vs substantial performance\\
3.Map consideration adequacy through benefit-detriment analysis\\
4.Prepare parol evidence rule applicability matrix\\

\begin{center}
\MakeUppercase{\romannum{2}}. Litigation Strategist
\end{center}
Responsibilities:\\
1.Develop FRCP-compliant pleading alternatives\\
2.Optimize discovery plan using proportionality standards\\
3.Calculate summary judgment probability scores\\
4.Prepare jury demand vs bench trial analysis\\

\begin{center}
\MakeUppercase{\romannum{3}}. 
Regulatory Compliance Auditor
\end{center}
Responsibilities:\\
1.Conduct Chevron/Mead framework analysis\\
2.Map agency guidance through FOIA-obtained materials\\
3.Prepare preemption challenge vulnerability index\\
4.Maintain regulatory change tracking dashboard\\
\end{tcolorbox}
\end{figure*}

\begin{figure*}[h]
\begin{tcolorbox}[title={Law Group of 3}, colback = cGrey, colframe = black,  coltitle=white,fonttitle=\bfseries\small, center title,fontupper=\small,fontlower=\small]
\begin{center}
\MakeUppercase{\romannum{1}}.  Solver
\end{center}
Responsibilities:\\
Analyze contract validity and compliance, evaluate breach of duty scenarios, and develop legal documentation frameworks.\\

\begin{center}
\MakeUppercase{\romannum{2}}. Critic
\end{center}
Responsibilities:\\
Audit regulatory adherence, identify compliance vulnerabilities, and verify proper application of legal precedents.\\

\begin{center}
\MakeUppercase{\romannum{3}}. 
Coordinator
\end{center}
Responsibilities:\\
Integrate litigation strategies with dispute resolution mechanisms, balance evidentiary requirements, and ensure procedural compliance.
\end{tcolorbox}
\end{figure*}

\begin{figure*}[h]
\section{Relevance Prompt}
\label{app:relevance prompt}
\begin{tcolorbox}[title={User Prompt}, colback = cGrey, colframe = black,  coltitle=white,fonttitle=\bfseries\small, center title,fontupper=\small,fontlower=\small]
You are an expert in identifying the domains of expertise required to solve a given problem. 
You will be provided with a question, and your task is to determine which domains from the following list are relevant: ['Math', 'Law', 'Business', 'Health']. \\
Please analyze the question and return the appropriate domains. 
There could be more than one domain that is necessary.\\
Please directly output a python list of the domains without other output.\\
Please limit your output to 2-3 domains.\\
For example: ['Med', 'Fina']\\
Please directly output the list that is loadable by python, no other output.                    
2-3 domains should be outputted, no more or less.
\end{tcolorbox}
\end{figure*}

\begin{figure*}
\section{All Experiments}
    \centering
    \includegraphics[width=\linewidth]{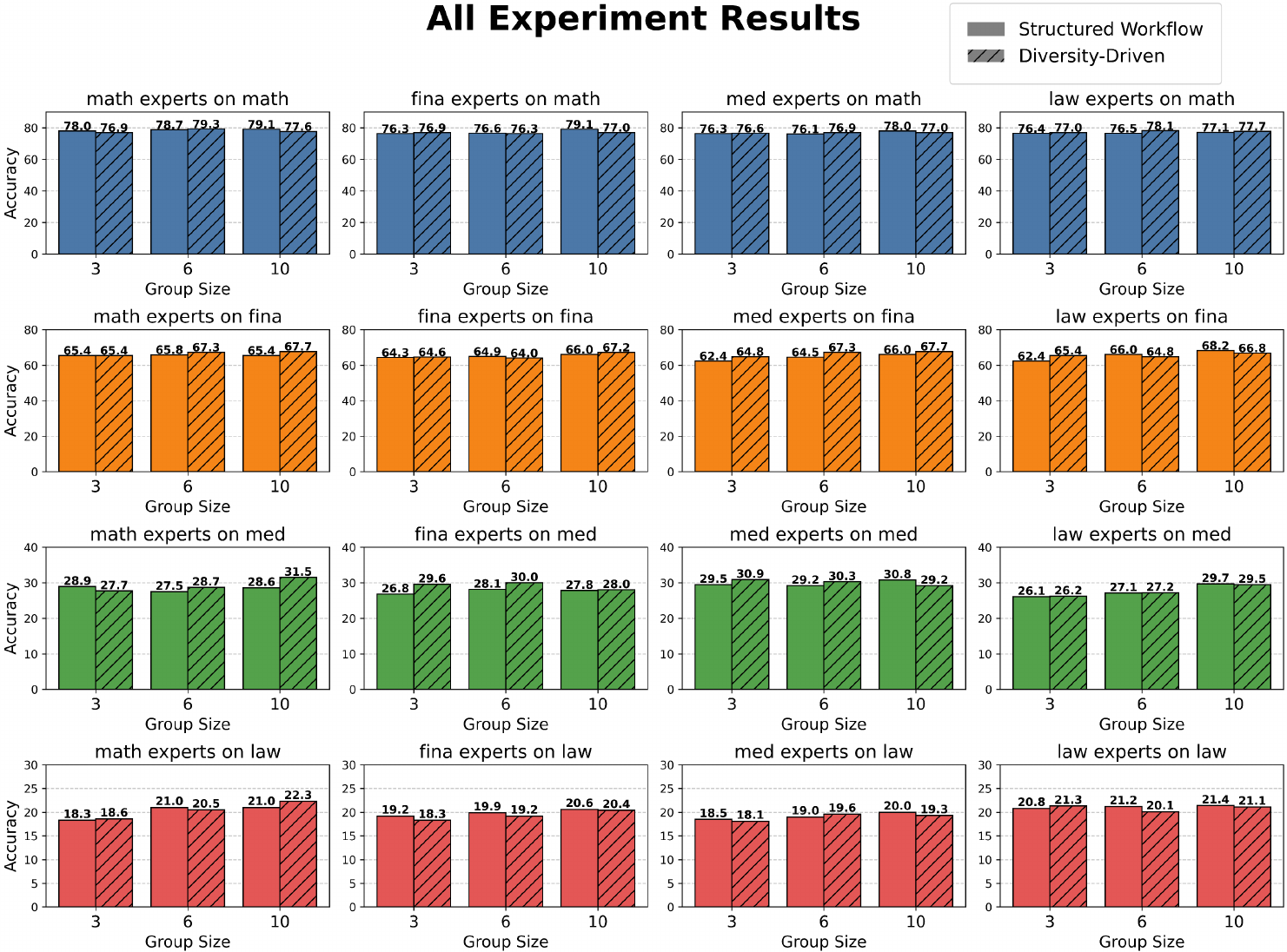}
    \caption{All Experiment Results Visualization}
    \label{fig:All_Exp}
\end{figure*}

\end{document}